\documentclass[runningheads]{llncs}
\usepackage{svg}
\usepackage{graphicx}
\usepackage{lscape}
\usepackage{longtable}
\usepackage{multirow}
\usepackage{booktabs}
\usepackage[square,sort,comma,numbers]{natbib}
\usepackage{subcaption}
\usepackage{multicol}
\usepackage{array}

\begin{document}
\title{State-of-the-Art Review: The Use of Digital Twins to Support Artificial Intelligence-Guided Predictive Maintenance\thanks{This work has been submitted to Springer for possible publication. Copyright may be transferred without notice, after which this version may no longer be accessible.} \thanks{This material is based upon work supported by the U.S. Army Research Office and the U.S. Army Futures Command under Contract No. W911NF-20-D-0002. The content of the information does not necessarily reflect the position or the policy of the government and no official endorsement should be inferred.}}
\titlerunning{SOTA: AI-Guided PMx DT}
%

\author{Sizhe Ma\inst{1}\orcidID{0000-0002-2532-5915} \and
Katherine A. Flanigan\inst{1}\orcidID{0000-0002-2454-5713} \and
Mario Berg\'es\inst{1}\orcidID{0000-0003-2948-9236} \thanks{Mario Berg\'es holds concurrent appointments at Carnegie Mellon University (CMU) and as an Amazon Scholar. This manuscript describes work at CMU and is not associated with Amazon.}}
\authorrunning{S. Ma et al.}
%
\institute{$^1$Civil and Environmental Engineering, Carnegie Mellon University, 5000 Forbes Ave, PA 15213, USA \\
\email{\{sizhem,kflaniga,mberges\}@andrew.cmu.edu}}
\authorrunning{S. Ma et al.}

\maketitle              
\begin{abstract}
In recent years, predictive maintenance (PMx) has risen to prominence, capturing significant attention for its potential to enhance efficiency, automation, accuracy, cost-effectiveness, and the reduction of human involvement in the maintenance process. Importantly, PMx has evolved in tandem with digital advancements, such as Big Data and the Internet of Things (IOT). These technological strides have enabled Artificial Intelligence (AI) to revolutionize PMx processes, with increasing capacities for real-time automation of monitoring, analysis, and prediction tasks. However, despite these leaps, PMx continues to grapple with challenges——ranging from poor explainability and sample inefficiency in data-driven methods to high complexity in physics-based models——that stymie its broader adoption. In this context, this paper posits that Digital Twins (DTs) hold the transformative potential to be seamlessly integrated into the PMx process as a means to surmount these limitations, thereby laying the groundwork for more extensive and context-sensitive automated PMx applications across varied stakeholders. Nevertheless, we contend that, at present, state-of-the-art DTs have not fully matured to the extent required to bridge these existing gaps. Our paper unfolds as a comprehensive roadmap for the evolution of DTs, geared towards redressing current limitations and fostering the large-scale progression of automated PMx. We structure our approach in three distinct stages: initially, we reference our prior work where we methodically identified and distinctly defined the Information Requirements (IRs) and Functional Requirements (FRs) for PMx, which serve as the fundamental blueprint for any ensuing unified framework. Subsequently, we embark on an exhaustive literature review across various disciplines to ascertain how DTs are presently being employed to integrate these previously identified IRs and FRs, and we also unveil extant standardized DT models and tools that stand to fortify automated PMx initiatives. Lastly, we spotlight the discernible gaps——in particular, those IRs and FRs that current DT implementations have yet to fully support—and meticulously delineate the components necessary for the realization of a comprehensive, automated PMx system. The paper culminates with a synthesis of our research directions, aimed at the seamless integration of DTs into the PMx paradigm to actualize this ambitious vision.
\end{abstract}
\section{Introduction}

Effective maintenance is vital for the efficiency, safety, and longevity of systems and their assets. Numerous strategies have been developed to enhance maintenance processes, including Predictive Maintenance (PMx), Corrective Maintenance (CM), Preventative Maintenance (PvM), and Total Productive Maintenance \cite{mobley2002introduction}. Among these, PMx stands out as the most appropriate for scenarios where avoiding run-to-failure is crucial, typical of many complex real-world systems. PMx is a forward-looking maintenance strategy that leverages data analysis, artificial intelligence (AI), and other advanced technologies to foresee potential system failures and implement preventative measures. This approach integrates real-time monitoring, data analytics, and historical data to anticipate equipment failures before they occur, enabling maintenance teams to plan repairs or replacements proactively. Previously, four primary levels of PMx have been identified \cite{haarman2017predictive}, each defined by distinct methodologies: \textit{Level 1:} visual inspection, \textit{Level 2:} instrument inspection, \textit{Level 3:} real-time condition monitoring, and \textit{Level 4:} continuous monitoring coupled with predictive techniques.

In recent years, PMx has experienced significant advancements, presenting new possibilities for improving system performance \cite{SAKIB2018267}. These developments have been driven by the adoption and integration of advanced technologies such as the Internet of Things (IoT), large-scale datasets, and increased computing power \cite{ran2019survey}. The IoT, a network of interconnected devices equipped with sensors, software, and internet connectivity, plays a pivotal role in this transformation. It facilitates seamless data collection and exchange between systems and the internet, reflecting the broader trend of technology integration in industrial processes and engineered systems. The expanded availability of data has created new opportunities for PMx, as more information can now inform maintenance decisions. Furthermore, PMx increasingly depends on complex algorithms and models capable of handling the vast amounts of data generated by IoT systems \cite{SAKIB2018267,serradilla2022deep}. With the advantage of growing computing resources, these algorithms support near real-time monitoring, analysis, and prediction of system behavior. Together, these technological advancements have led to more accurate predictions and have enhanced the standard of maintenance decision-making capabilities.

Despite the significant progress in PMx, various challenges remain that hinder its full development and broader adoption. These challenges arise from the complexities inherent in both data-driven and physics-based modeling techniques crucial to PMx. Data-driven PMx encounters difficulties such as the scarcity of failure data, as systems rarely operate until failure, complicating accurate failure prediction \cite{alimohammadi2022predict,wen2022recent}. Moreover, current models often lack explainability, leading to skepticism about the accuracy of algorithms and diminishing trust in the models' decisions and recommendations \cite{shukla2020opportunities}. Additionally, many PMx algorithms are not sample-efficient, requiring extensive datasets to produce reliable results, which is problematic given the infrequency of machine failures \cite{wen2022recent}. Conversely, physics-based modeling techniques are characterized by their intractable complexity, making them resource-intensive and demanding considerable expertise to develop \cite{aivaliotis2019methodology}. These models also tend to have limited generalizability, as they are typically tailored to specific types of equipment \cite{boje2020towards}. To address these limitations, hybrid approaches such as Physics-Informed Machine Learning (PIML), Scientific Machine Learning (Sci-ML), and integrated hybrid models have emerged as promising solutions \cite{sepe2021physics,nascimento2019fleet}. However, these methods, while effectively leveraging sensor data, often incorporate only a portion of available physical knowledge into the modeling process. For instance, PIML embeds physical rules as implicit constraints within ANN models without comprehensive integration, underscoring the evolving nature of these approaches and the necessity for further research and development.


The current manuscript takes this discussion further. It seeks to build upon the foundations established by our previous research \cite{ma2023twin}, contextualizing each requirement with tangible examples, exploring state-of-the-art literature individually, and gleaning insights from industry standards and expert perspectives. In this context, the present paper builds upon that foundation without redundantly restating the selection procedure of manuscripts; instead, it focuses on the identification of critical elements that constitute a computational framework of DT. We delve into standardized DT models and environments that can facilitate the integration and implementation of PMx strategies. While we argue that DTs have this transformative potential within the PMx space, state-of-the-art DTs have not yet achieved a level of maturity and robustness that is requisite for widespread industrial adoption. So, we further assess the current status of individual manuscripts in terms of their fulfillment of the identified IRs and FRs, thereby providing a more detailed and granular understanding of existing gaps. This paper serves as a roadmap for DT developments that will address existing shortcomings and support the advancement of automated PMx at scale.

A systematic approach comprising three primary stages, which is designed to underscore the interaction between PMx and DT. We begin in Section~\ref{sec:IRFR} with a presentation of the identified IRs and FRs from our previous work. These IRs and FRs form the foundation for any unified PMx DT. We then move to Section~\ref{sec:DTT}, where a comprehensive literature review is conducted to explore the current utilization of DTs in incorporating these IRs and FRs. This section also details the supporting DT models and environments, highlighting the existing standardized DT tools that facilitate automated PMx tasks. The subsequent section, Section~\ref{sec:GCO}, serves to connect the concepts discussed in the previous sections. Here, we identify and scope the gaps, particularly those IRs and FRs not currently supported by DTs. This critical analysis helps to pinpoint the missing elements necessary for fully automated PMx. Finally, the chapter concludes in Section~\ref{sec:Conclusions}, where we summarize the research agenda for expanding and integrating DTs into the PMx process. This concluding section aims to chart a path forward for realizing the vision of seamlessly blending PMx with DT technologies, thereby enhancing the efficiency and efficacy of maintenance strategies.

\section{Predictive Maintenance, Informational Requirements and Functional Requirements} 
\label{sec:IRFR}
\subsection{Predictive Maintenance and Related Strategies}
\label{subsec:module}

Maintenance practices play a pivotal role in ensuring the safe, prolonged, and economically viable use of equipment across diverse sectors. Over the past two decades, the field has witnessed remarkable advancements, refining and enhancing the way industries approach equipment upkeep. At the core of these advancements lie three predominant maintenance strategies: CM, PvM, and PMx.

\begin{figure}[h]
\centering
\begin{subfigure}{.5\textwidth}
  \centering
  \includegraphics[scale=0.45]{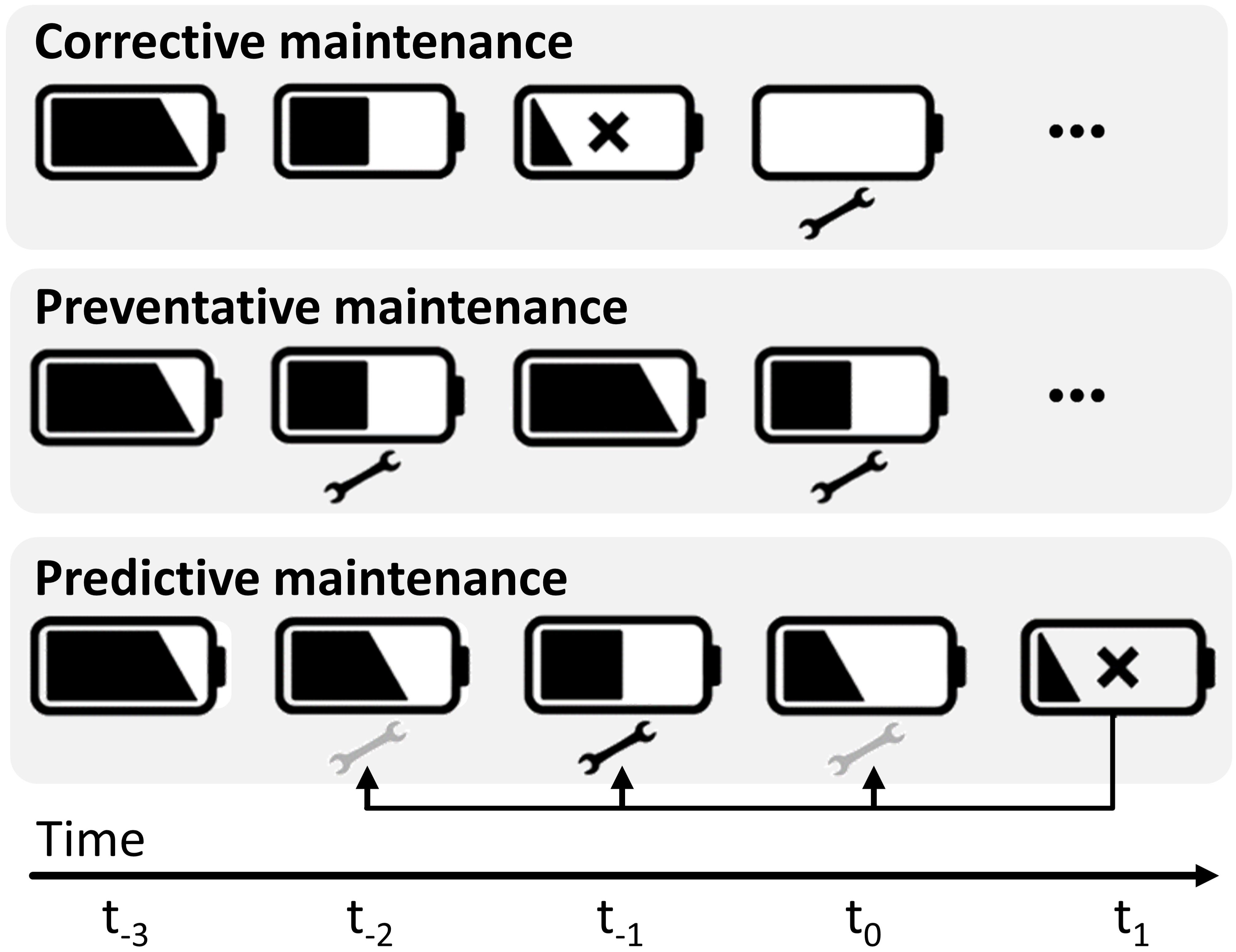}
  \caption{}
  \label{fig:pmxa}
\end{subfigure}%
\begin{subfigure}{.5\textwidth}
  \centering
  \includegraphics[scale=0.47]{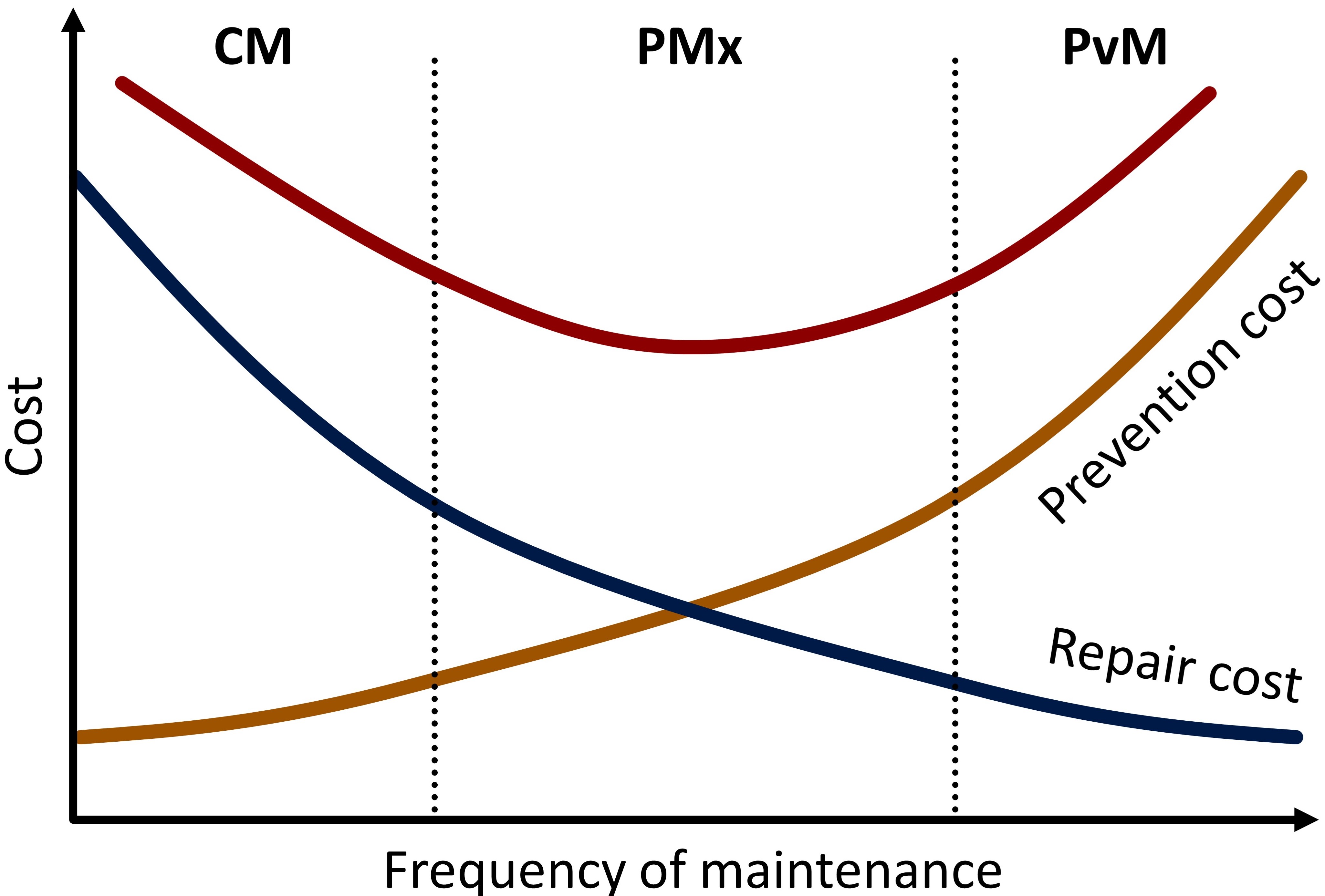}
  \caption{}
  \label{fig:pmxb}
\end{subfigure}
\caption{Comparison between different maintenance techniques with respect to (a) the underlying mechanisms and (b) the economy \cite{ran2019survey}.}
\label{fig:pmx}
\end{figure}

Delving deeper, as depicted in Figure~\ref{fig:pmxa}:
\begin{itemize}
    \item CM is reactive in nature, instigating maintenance actions only after a component or system exhibits a fault. While it addresses immediate issues, this approach can lead to unforeseen breakdowns.

    \item PvM operates on a calendar-based system, scheduling routine maintenance checks and repairs. This proactive approach aims to preemptively mitigate potential failures, but it may also inadvertently result in redundant checks and wastage of resources.

    \item PMx is the most advanced among the three, predicting equipment failure by estimating its forthcoming states. It recommends specific maintenance tasks from a broad spectrum of choices, considering both resource accessibility and economic implications.
\end{itemize}

Further elucidation on the economic implications of these strategies is provided in Figure~\ref{fig:pmxb}. The visualization highlights the comparison of the cost-effectiveness of reactive repairs against preventive measures. Observably, while CM might grapple with soaring replacement expenditures and secondary damages, PvM could inadvertently lead to superfluous repairs and resource depletion. PMx masterfully bridges the gap, determining the most opportune moments for maintenance interventions. Any divergence from this meticulously crafted schedule could tip the balance, either causing resource wastage or amplifying the risk of collateral damage \cite{jimenez2019system}.

\subsection{Informational Requirements and Functional Requirements of Predictive Maintenance} \label{subsec:IRFR}
\begin{figure}[h]
    \centering
    \includegraphics[scale=0.32]{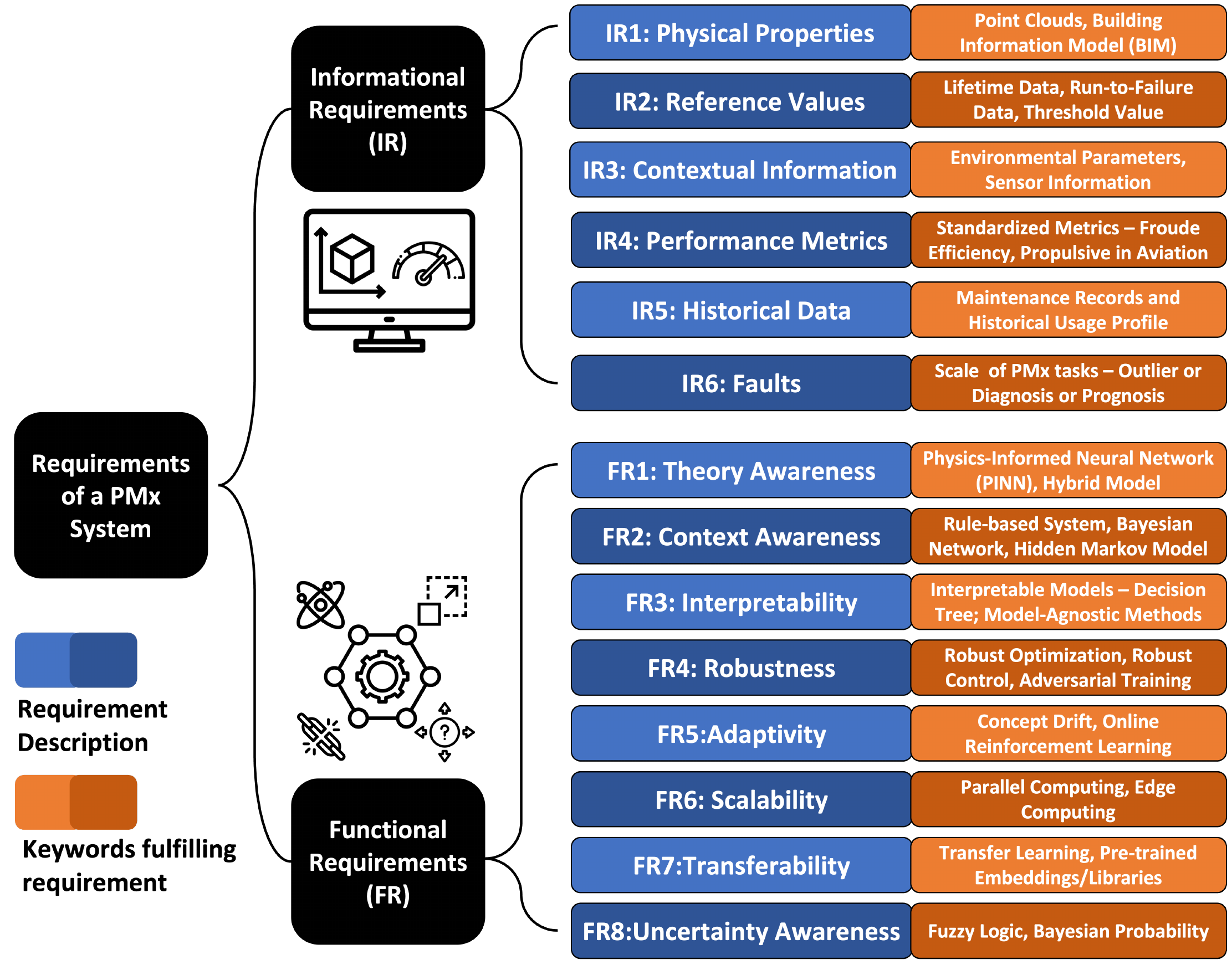}
    \caption{Overview of the identified IRs and FRs.}
    \label{fig:IRFR}
\end{figure}

Traditionally, PMx harnesses granular insights about the system, such as real-time operational parameters and configurations, to finetune its maintenance agenda. This optimization predominantly relies on prognostic techniques to identify future usage profiles and the RUL of the equipment. With the contemporary surge in computational prowess, ML algorithms have led to significant advancements in AI-guided optimization for PMx. Their evolution has paved the way for groundbreaking strides in AI-driven optimization techniques tailored for PMx. Building on this foundation, it is pivotal to revisit the IRs and FRs we delineated in our prior research \cite{ma2023twin}. Following the step of system and software industry \cite{sommerville2009deriving, lana2021data}, these systematically identified IRs and FRs provide a structured framework and critical blueprint, enabling the effective implementation and optimization of PMx strategies within the evolving landscape of AI and digital technology, as shown in Figure~\ref{fig:IRFR}.
\newline

IRs:
\begin{enumerate}
    \item Physical Properties: The spatial information and characteristics of an asset or its components.
    \item Reference Values: The benchmark for an asset or its components to fail.
    \item Contextual Information: The data used to determine the current or future states of a component or to make decisions and recommendations without being directly derived from the physical entity being maintained. 
    \item Performance Metrics: The metrics that quantify the performance of operations of an asset or its components.
    \item Historical Data: The maintenance records and usage profiles of an asset or its components.
    \item Faults: The anomalies, malfunctions, or failures of an asset or its components.
\end{enumerate}

FRs:

\begin{enumerate}
    \item Theory Awareness: The ability of the system to maintain efficient and consistent representations of the physical phenomena involved in the assets' operation \cite{lana2021data}.
    \item Context Awareness: The ability of the system to perceive and adapt to various operational and environmental factors \cite{GALAR2015137}.
    \item Interpretability: The ability of the system to generate human-interpretable outputs \cite{vollert2021interpretable}. 
    \item Robustness: The ability of the system to maintain acceptable performance under potential disturbances in both physical and digital domains \cite{lana2021data}.
    \item Adaptivity: The ability of the system to modify its internal process or behaviors based on the deterioration or evolution of an asset \cite{lana2021data}.
    \item Scalability: The ability of the system to maintain performance across a diverse range of workloads or to be extended to varying scales.
    \item Transferability: The system's capability to uphold its performance when deployed on assets or conditions distinct from those on which it was initially trained \cite{lana2021data}.
    \item Uncertainty Awareness: The system's ability to recognize and quantify uncertainty inherent in its input, process modeled, and/or outputs.
\end{enumerate}

\section{Digital Twins and Threads}
\label{sec:DTT}
\subsection{Digital Twins}
\label{DTD}
\begin{figure}[h]
    \centering
    \includegraphics[scale=0.2]{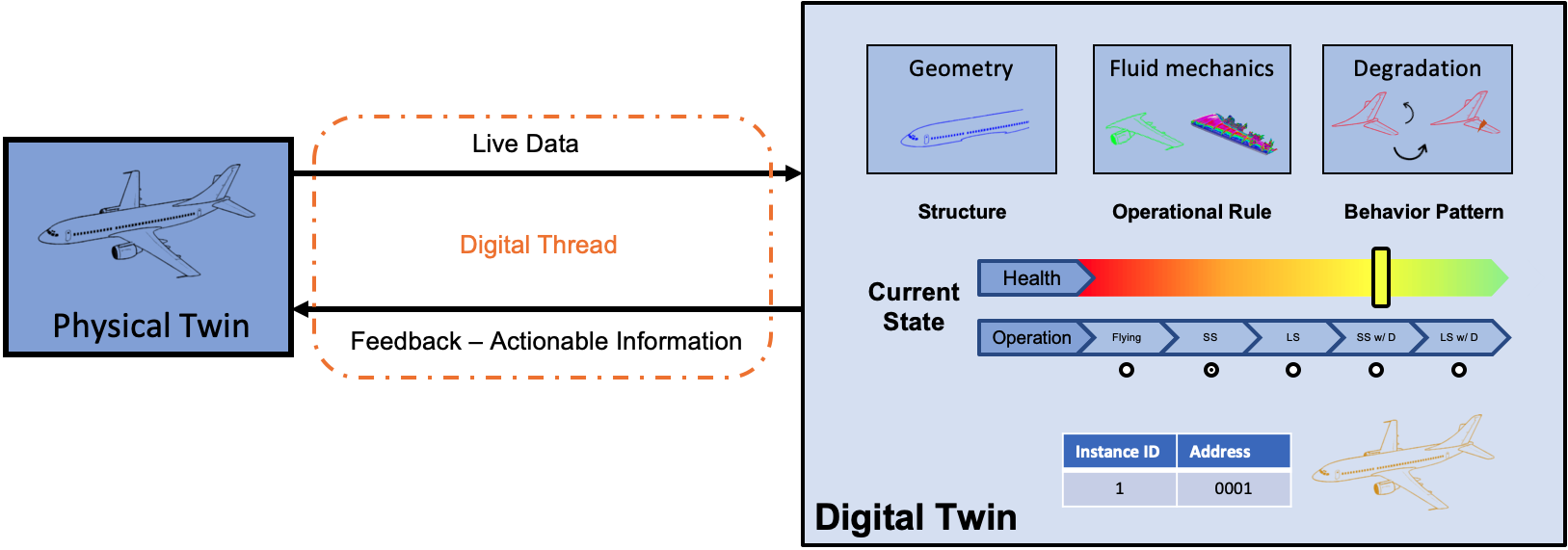}
    \caption{Digital Twin and Digital Thread used for Aviation-related PMx.}
    \label{fig:DT}
\end{figure}

The concept of Digital Twins (DTs) initially emerged without a widely agreed-upon definition, as various fields adapted the term to fit their unique applications. The term gained a more defined structure when Grieves offered a comprehensive description, defining DTs as \textit{``a set of virtual information constructs that fully describes a potential or actual physical manufactured product from the micro atomic level to the macro geometrical level"} \cite{grieves2017digital}. Figure \ref{fig:DT} shows how these elements operate with each other visually in the context of PMx.

Grieves' seminal work further introduces two critical applications of DTs, labeled as \textit{`Possibilities.'} The first concept focuses on integrating usage data into the design phase, and the second, known as \textit{``front-running" a simulation in real-time}, aims to prevent unpredictable and undesirable outcomes. These concepts have evolved into key elements of DT technology, commonly referred to in academia and industry as `design-in-use' and `what-if simulation.'

The evolution of these concepts highlights not just the increasing relevance of DT technology but also the diverse ways in which DTs have been developed and applied across various sectors. In the last five years, different industries have adopted these visions to suit their needs. The aerospace and automotive industries, for instance, have largely leveraged the `what-if simulation' aspect to meet their complex systems and safety requirements \cite{ezhilarasu2019management}. The manufacturing sector, conversely, has leaned towards the `design-in-use' model, using real-time operational data to enhance manufacturing processes and product designs \cite{HONGLIM202289}. The Medical and Healthcare sectors have shown a balanced application of both models, employing `what-if simulations' for medical procedures and `design-in-use' for optimizing medical equipment and systems \cite{HALEEM202328}.

These two primary characteristics of DTs exemplify how their definitions and functionalities are shaped by the varied expectations of stakeholders in different fields. As new technologies emerge, the scope and potential of DTs have continued to grow, incorporating features like advanced visualization, interoperability, and Human-System Interaction (HSI). This expansion leads to an even greater divergence in the understanding and application of DTs.

\subsection{Digital Threads}
The term Digital Thread (DThread), in a PMx context, refers to the interconnected flow (wired or wireless) and systematic integration of data across the lifecycle of an asset or a system \cite{kunzer2022digital}, as shown in Figure \ref{fig:DT}. This encompasses the data generated from the design phase, operation phase, and maintenance phase, facilitating a comprehensive understanding of the asset's health and performance. Conceptualizing DThreads can be aided by envisioning them as links or threads, which weave together information such as sensor data, operational data, maintenance records, inspection data, etc. Analogous to sewing threads, the continuity of DThreads is essential, as any break in these connections obstructs the seamless flow of data.

It is vital to distinguish between DThread and DT to avoid any misconception that may occur from mixing the two terms. DTs focus on creating the virtual replica of a product that can mimic the behavior of the physical asset with the help of high-fidelity simulations or data-driven methods, whereas DThreads provide actionable information flow between physical and virtual assets bidirectionally. Thus, DThread can be interpreted as the link including all the required information to generate and supply updates to a DT \cite{singh2018engineering}. It is worth highlighting that DThread is neither DT nor part of DT, but rather an essential element of DTF to realize DT. DT heavily relies on DThread to access and integrate data from various sources and stages along the physical entities' lifecycle. Without a continuous flow of information, DT could not be kept up-to-date accurately, thus merely a simulation model. While DThread is a prerequisite for DT, the opposite does not hold, indicating that DThread can be operationalized independently of a DTF \cite{hedberg2020using}. DThread is used synonymously with connected supply chain \cite{kunzer2022digital} and simplified information-relay frameworks \cite{margaria2019digital}. However, it is undeniable that, in the absence of a virtual counterpart to the physical entity, the information flow provided by DThread could not be effectively integrated for high-level downstream tasks \cite{singh2021twin}, such as optimization, and decision-making.

\subsection{Digital Twin Frameworks} \label{sec:dtf}
There have been multiple attempts at proposing a DTF as the solution for PMx tasks \cite{onaji2022frame,sana2022frame}. Depending on the assets they are considering for maintenance and the PMx tasks applied, as well as the algorithms used to model the physical twin, the DTF implemented varies significantly. In order not to mislead or confuse the reader, we first present minimal viable components that a DTF should have so as to pave a roadmap for further investigating the mapping between PMx and DT on both general and individual levels. This list of components is inspired by \cite{kunzer2022digital} and expanded by selected literature after the three-step process \cite{ma2023twin}. In the meantime, an accompanying contextual example (\textit{italics}) is provided in parallel. The example considers creating a DT for the helicopter to aid the PMx process aimed at maintaining the health of a helicopter's tail rotor drive shaft (TRDS) towards alignment and imbalance. DTFs from selected literature only have a subset of components listed below, which has the potential to realize IRs and FRs:

\begin{itemize}
    \item Physical Twin (PT) - PTs are practical artifacts that could potentially cover a wide range of applications (e.g., system, product, equipment). Without PT, the DT is a conventional simulation model. Example: \textit{The TRDS itself.}
    \item Physical Twin Environment (PTE) - Depending on the PT, PTE exists as a set of measurable variables in which the PT lives. It provides contextual information. Without PTE, the DT is an incomplete representation of the PT. Example: \textit{The surrounding conditions that the helicopter operates in, including the atmospheric pressure, humidity, and temperature.}
    \item Digital Twin (DT) - A digital representation of the physical asset generated by computers. From a functionality point of view, it could be viewed as an integrated model that combines information models (e.g., geometry, material configurations) and simulation engines (e.g., static models, dynamic models), etc. From a solution point of view, it could utilize physics-based models, data-driven models, or even expert-based models. Without DT, the PT lacks an integrated information source to exchange data and perform analysis. Example: \textit{The integration of information models like the geometry, operating conditions, and health conditions (e.g., alignment, imbalance) of TRDS, and dynamics model like finite element model based on vibration analysis.}
    \item Digital Twin Environment (DTE) - The `virtual world' in which the DT exists. The practical use could be replicating PTE for different simulation scenarios. Without DTE, what-if analyses are narrow, if not absent. Example: \textit{The Digital representation (a matrix in its simplest form) of atmospheric pressure, humidity, and temperature.}
    \item Digital Environment (DE) - The platforms that sustain the functionality of DTs. The actual selection depends on the exact rule and behavior to be modeled. Without DE, the DT is merely a concept that remains unimplemented. Example: \textit{Ansys Fluent\footnote{Ansys Mechanical [website], www.ansys.com/products/structures/ansys-mechanical, (last accessed October 2022).}, due to its interoperability and scriptability.}
    \item Instrumentation - The sensors, detectors, and other measurement tools, including digitized manual inspection notes, periodically collect data from the physical world. They are segments independent from PT or PTE that provides DT with information/knowledge of the two. Without instrumentation, there is no data from PT to transmit to DT. Example: \textit{Accelerometers on each drive shaft segment and meteorological sensors on the helicopter.}
    \item Realization - The actuators or human resources, depending on the PT, periodically realize actions/decisions generated from the digital world. They are segments independent from PT or PTE that receives and actualize information/knowledge generalized by the Analysis component. Without realization, there is no action/decision from DT to be actualized on PT. Example: \textit{Human resources that realize the action generated by the system, such as inspection or repair.}
    \item Digital Thread (DThread) - The digital connection between PT and DT allows  bi-directional information flow between two twins. Without DThread, there is no data exchange between different components in the DTF, and the DT is no better than an analog model. Example: \textit{Data acquisition and management software like LabVIEW\footnote{LabVIEW [website], www.ni.com/en-us/shop/labview.html, (last accessed October 2022).} plus IoT platforms or middleware like Microsoft Azure IoT Suite\footnote{Microsoft Azure IoT Suite [website], azure.microsoft.com/en-us/products/iot-hub, (last accessed November 2022).}.}
    \item Historical Repository (HR) - The storage system for the past usage patterns and maintenance records of the PT. Unlike other components that form part of the DT, HR exists as a separate entity. It captures and retains extensive historical data that are not immediately discernible from examining the PT. A pointer within the DT is created to indicate the location of the HR pertinent to a specific PT instance, allowing the HR to maintain an intimate connection with the DT while retaining its distinct identity. Without HR, there is no longitudinal analysis, and therefore, hinders the effective prediction of future performance and potential issues. Example: \textit{Time-series database like InfluxDB\footnote{InfluxDB [website], www.influxdata.com, (last accessed November 2022).} stores in a hard drive or cloud-based solution.}
    \item Analysis Module - The module that analyses information from DT and generates alarms, recommendations, decisions, or actions in the PMx context. In practical implementation, observe–orient–decide–act (OODA) could be viewed as a representation of the analysis module. Without the analysis module, the DT simply mimics the PT and is capable of simulation without generating actionable information. Example: \textit{The OODA process for this example would unfold as follows. First, the current state of the system, with regard to its health and operation, is observed through a user interface. Next, the observed states are oriented in relation to reference standards. A decision is then made as to whether an inspection or repair is needed based on the orientation stage. Finally, appropriate actions are taken in response to the prior decision.}
    \item Accountability Module - This component furnishes justifications for the alerts, recommendations, decisions, or actions delivered by the Analysis module. Periodic human review is required to ensure the validity of these outputs and to assess whether the Digital Twin or Analysis module necessitates updates or retraining. Without the explanation module, attributing responsibility (accountability) for unfavorable or even damaging results from the system's output becomes ambiguous. This lack of clarity can significantly undermine user trust in the framework, thereby hindering its successful deployment. Example: \textit{SHAP method that elucidates the significance of vibrations at various locations in the prediction of misalignment and imbalance.}
    \item Query \& Response (QR) - QR accordingly accesses and gathers information from the digital twin side. Without the functionality of QR, DT is merely an integration of unorganized information without efficient ways to interact with users. Example: \textit{Ansys Mechanical’s API that allows user to acquire information or modify the settings of DT.}
\end{itemize}

\subsection{Standardized Models in Digital Twin and Standardized Digital Environments} \label{m&e}
Following the list of critical components within DTF, the ensuing section consolidates and discusses the standard models and environments observed in the literature, providing valuable insight into the foundational elements of the DT concept. When certain models or environments are identified, examples from unselected literature are also mentioned to offer readers more insights when implementing their own DT while understanding the mapping between PMx and DT. These discussions focus primarily on two key questions: what kind of models can serve as a part of the DT, and which DE can effectively support these models. The significance of discussing models in DT lies in their ability to capture and simulate the complexities of physical systems. The choice of the model directly impacts the accuracy and performance of the DTs and PMx tasks. Different scenarios and systems may require different kinds of models, be they physics-based, data-driven, or even expert-based models. By examining and categorizing the models, we offer a comprehensive view of the current landscape, informing future design and development of DT in various PMx-related domains. The importance of examining DE is underscored by their integral role in enabling the operation and interaction of DTs. DEs are the computational platforms or ecosystems that underpin the functionality of DTs. They can range from general-purpose computing platforms, such as Amazon Web Services (AWS)\footnote{Amazon Web Services [website], https://aws.amazon.com/, (last accessed November 2022).}, to dedicated software solutions, such as Ansys Twin Builder\footnote{Ansys Twin Builder [website], www.ansys.com/products/digital-twin/ansys-twin-builder, (last accessed November 2022).}, designed specifically to support complex simulations, data analysis, and real-time interactions between the PTs and DTs. While general-purpose computing platforms provide flexibility and broad capabilities, dedicated software solutions can offer more immediate, out-of-the-box value, and efficiency for the development and deployment of DTs. Given the scale of PMx DT, only dedicated software solutions are discussed in the following section.

\subsubsection{Models in Digital Twin} \label{models}
The models in DT are the models operating within the virtual space to represent certain characteristics or properties of their PT. Depending on the motivation and purpose of the task, the models being used can be significantly different. However, thanks to the increase in computing resources, one thing in common for most DTs in literature is the integration of multiple models focusing on different aspects of the physical twins. There were multiple attempts to categorize these sub-models or elements that could potentially be part of the DT \cite{liu2021biomimicry,qi2021smart}. Before discussing actual models, we first present a non-exhaustive classification of these elements and their projections to IRs and FRs in Table \ref{modelirfr}. This act is crucial, because a non-exhaustive list, while it may not capture every conceivable element, provides a valuable reference and framework for understanding the breadth and complexity of the models that may be included within a DT. This list serves as a foundational layer to start the exploration of DT's architecture and the various ways it can be adapted to suit different tasks or objectives. Moreover, the classification of these elements and their corresponding IRs and FRs enables a systematic analysis of how different elements can contribute to the overall functionality of PMx DTs. It provides a roadmap for understanding the interaction between these elements and their influence on the system's performance. In the meantime, accompanying contextual examples (\textit{italics}) are provided in parallel under the same example mentioned in Section \ref{sec:dtf}:
\begin{itemize}
    \item Information Model: This model defines the geometric, topological and operational information of the physical entity (e.g., dimension, configuration). Example: \textit{The computer-aided design (CAD) file of the helicopter or its TRDS.}
    \item Static Rule Model: This type of model defines the physical rules/properties that define the static (not time-dependent) input-output behavior of the asset or component. These models \textbf{only} consider the static rule of the physical entity instead of the dynamic evolution/deterioration. Example:  \textit{The finite element model, representing the TRDS, ingests data from the load sensor and consequently generates strain outputs.}
    \item Dynamic Behavior Model: Like the preceding type, this model also defines input-output behavior but as it relates to time. These models could include in scope the evolution/deterioration under ideal circumstances or disturbances. Behavior models are an essential factor for the simulation ability of DTs. Example:  \textit{The Hidden Markov model that models the change in the health state of the TRDS over time.}
    \item Tools for Integration: It describes the process that determines the interactions between the aforementioned models, which plays a pivotal role in the DTs, as it guarantees the functionality of the overall system. Example:  \textit{Snippets of code that define, based on the different operating status of the helicopter and system (train or validation), how the outputs from the finite element model are transferred to the Hidden Markov model.}
\end{itemize}

\begin{table}[htbp]
\caption{Mapping between different types of models and requirements.}
\begin{center}
\begin{tabular}{p{0.18\linewidth}|p{0.18\linewidth}p{0.18\linewidth}p{0.18\linewidth}>{\centering\arraybackslash}p{0.18\linewidth}}
\hline
    \centering \textbf{Requirement Identifier} & \centering \textbf{information Model} & \centering \textbf{Static Rule Model} & \centering \textbf{Dynamic Behavior Model} &  \textbf{Tools for integration}\\
    \hline
    \centering IR1 & \centering x & \centering x & \centering x & \\\hline
    \centering IR2 & \centering x & \centering x & \centering x & \\\hline
    \centering IR3 & \centering x & & & \\\hline
    \centering IR4 & \centering x & \centering x & \centering x & \\\hline
    \centering IR5 & \centering x & & \centering x & \\\hline
    \centering IR6 & \centering x & & & \\\hline
    \centering FR1 & & \centering x & \centering x & \\\hline
    \centering FR2 & \centering x & & & x\\\hline
    \centering FR3 & & \centering x & \centering x & \\\hline
    \centering FR4 & & \centering x & \centering x & x\\\hline
    \centering FR5 & & \centering x & \centering x & x\\\hline
    \centering FR6 & \centering x & & \centering x & x\\\hline
    \centering FR7 & & \centering x & \centering x & x\\\hline
    \centering FR8 & & \centering x & \centering x & \\\hline

\end{tabular}
\label{modelirfr}
\end{center}
\end{table}

The remaining section explores DT models, their strengths, and their limitations individually. Examples are incorporated with each model to offer readers more channels for extensive reading, as all examples lie within the intersection of DT and PMx.\\
\\
\indent \textbf{Point Cloud Modeling - Information Model:} Point clouds are discrete data sets representing 3D shapes or objects in space. They typically encompass Cartesian coordinates, electrical responses, and asset textures. It is typically used to capture larger, complex, and spatially distributed assets in PMx, such as industrial equipment and facilities \cite{stojanovic2018towards,sommer2019automatic} and transportation assets \cite{xue2020lidar}. This typically involves the output of 3D scanners \cite{zhang2022highly} or photogrammetry software \cite{mohammadi2021quality} during the inspection or monitoring phase, which often takes the form of point clouds. The point cloud data and the derived 3D model are used in various level of PMx tasks, including anormaly detection and diagnosis \cite{wu2023high,wang2022digital}. As critical components for capturing information about physical entities, point clouds serve as a foundational modeling technique used for information modeling. Point cloud modeling techniques can be classified based on various standards, including the learning mechanism. For example, according to the learning mechanism, point cloud modeling techniques could be grouped into categories such as non-learning, supervised learning, unsupervised learning, and reinforcement learning \cite{xue2020lidar}. One of the key strengths of point cloud modeling lies in its high resolution. Furthermore, the raw data necessary for point cloud modeling is generally easy to collect. However, working with 3D point clouds also poses certain challenges. For instance, raw point clouds demand data mining techniques to transform them into an efficient and useful format suitable for point cloud modeling \cite{macher2017point}. Moreover, the practicality of the point cloud modeling process is a critical concern. It must not only be less time-consuming but also offer higher accuracy than manual processes \cite{sommer2019automatic}. Thus, while point cloud modeling is a powerful tool for capturing and representing physical entities, it is not without its challenges, especially when applied in the context of PMx DT. \\

\textbf{Building Information Modeling (BIM) - Information Model:} Regarded as a platform for generating precise and interoperable information models, BIM effectively represents the physical properties of a given structure. It is typically used in the context of PMx for built environments or structures, like buildings, infrastructures, and even cities \cite{khajavi2019bim,lu2020bim}. BIM models are typically created during the design phase and further updated throughout construction phase of a building or infrastructure project, making the models created via BIM typically exhibit richer semantics and a more organized structure. As BIM supports a lifecycle perspective, enabling users to monitor and maintain an asset from its inception through to its decommissioning, it could assist on a wider range of PMx tasks than point clouds, such as decision-making and prognosis \cite{jiang2021digital,hichri2013point}. However, the role BIM plays within DTF is relatively limited, primarily due to its lack of data manipulation or predictive capabilities \cite{pan2021bim}. Another challenge associated with BIM is related to entities that lack pre-existing BIM documentation. In such cases, the effort required to create and update a BIM model is often disproportionate to the resultant output. This imbalance can present significant obstacles to the practical implementation of BIM in these contexts \cite{khajavi2019bim,arayici2008bim}. As a result, the potential of BIM in the field of PMx DT may be hindered by these inherent limitations.\\

\textbf{Finite Element Method (FEM) - Rule Model and Behavior Model:} FEM represents a crucial branch within the sphere of Computer-Aided Engineering (CAE). It is especially noteworthy for its process of discretization, which simplifies complex systems into manageable elements. This procedure is particularly useful when dealing with irregular geometric shapes, heterogeneous material properties, and complex boundary conditions, all of which often prove unmanageable for analytical methods \cite{zhang2019fast}. Rather than deriving and solving equations as a whole, the components are divided into smaller parts, each governed by simpler equations. The solutions to these are subsequently combined to yield the final results (e.g., dislocation in structural analysis, temperature in heat transfer analysis, etc.). This makes FEM an ideal candidate for both rule and behavior modeling of intricate structures in the context of PMx DT. In static problems, FEM is used to determine the state of the system under a specific set of conditions at a particular point in time \cite{kapteyn2022data}. In dynamic problems, FEM can also be used when the state of the system changes over time due to forces, heat transfer, etc. In these cases, FEM is used to determine the relationship between states at different points in time \cite{urbas2021machine}. FEM is commonly applied to a wide variety of assets, such as structural components \cite{kapteyn2020toward} and mechanical systems \cite{tuegel2011reengineering}. Similar to CFD, the creation of the model typically occurs during the design phase and operational phase. Particularly in the operational phase to aid maintenance, The FEM could be used in various levels of PMx tasks, including anomaly detection, diagnosis, and prognosis \cite{tuegel2011reengineering}. However, like many other CAE approaches, FEM is not without its challenges. Notably, it grapples with complexity when addressing sophisticated systems \cite{kapteyn2022data}. To tackle this issue, reduced-order modeling is frequently employed in tandem with FEM in practical DT contexts. This combination allows for the fulfillment of real-time operation requirements, a crucial aspect of effective PMx strategies \cite{hurkamp2020combining, kapteyn2022data}. However, the application of such reduced-order models requires careful consideration of their validity range and potential inaccuracies. As such, a thorough understanding of the system and its nuances is essential for the successful implementation of FEM in the PMx DT context.\\

\textbf{Fuzzy Logic Modeling - Rule Model and Behavior Model:} The application of DT, especially within the context of PMx, is substantially driven by the necessity to generate precise and adaptive input-output projections from the data. Catering to this requirement, fuzzy logic modeling has been leveraged as the static rule and dynamic behavior model of DT, due in large part to its capabilities as a universal approximator \cite{alves2021digital}. Fuzzy logic modeling is frequently applied in PMx scenarios involving assets that have complex or non-linear behavior, uncertainty, imprecision, or where data is ambiguous or incomplete, such as automotive systems \cite{venkatesan2019health}. In the context of PMx, experts build a fuzzy logic model based on the understanding of the system and its potential failure modes during the design phase. The model is then updated during the operational phase and used primarily during the maintenance phase. Due to the lack of temporal information and confidence interval, fuzzy logic modeling is mainly used for PMx tasks such as anomaly detection and fault diagnosis. Beyond its approximation abilities, fuzzy logic also offers the interpretability of input-output projections. This characteristic fulfills a vital functional requirement mentioned earlier in Section~\ref{subsec:IRFR} \cite{ANTONELLI2016649}. Consequently, the interpretability of fuzzy logic models aids in understanding the reasoning behind specific outputs, providing crucial insights for decision-making. Despite its many benefits, the scalability of fuzzy logic models may pose a limitation to their practical application. As the number of inputs, outputs, and fuzzy sets grows, there is a corresponding exponential growth in the number of rules and membership functions within a fuzzy logic model \cite{alves2021digital}. This steep increase can complicate the model's application, making it a potential barrier to scalability in real-world settings.\\

\textbf{Computational Fluid Dynamics (CFD) Modeling - Behavior Model:} CFD modeling constitutes another significant branch within the realm of Computer-Aided Engineering (CAE). While the use of FEM may occasionally be part of CFD modeling, commercial CFD packages typically employ Finite Volume Methods (FVM), as the convection term in fluid dynamics proves too computationally demanding for FEM to handle effectively. Consequently, the forthcoming discussion on CFD primarily revolves around the use of FVM. CFD is specifically dedicated to analyzing and solving issues related to fluid flows through the application of numerical analysis and data structures. The variables within the fluid flow, such as mass and momentum, adhere to certain differential equations that define dynamic behaviors, making CFD an indispensable modeling technique for DTs involving fluid dynamics \cite{phanden2021review}. In the context of PMx, fluid dynamics mean assets such as turbines, pipelines and pumps, and HVAC systems \cite{deng2021bim}. The creation of this dynamic behavior model typically occurs during the design phase to support the design process and is subsequently utilized during the operational phase to assist with maintenance tasks. The CFD modeling could be used in various levels of PMx tasks, including anomaly detection, diagnosis, and prognosis. However, CFD modeling, like other CAE applications, is not without its drawbacks. The primary challenge resides in the complexity involved in simulations using differential equations, which often renders real-time responses unachievable. This issue is particularly pronounced in the context of PMx DT, where timely information and responses are essential. In practice, these challenges are frequently mitigated by integrating ML techniques and reduced-order modeling. These methods effectively cover a broad spectrum of operating conditions, bypassing the need for brute-force simulations \cite{molinaro2021embedding,aversano2019application}.\\

\textbf{Bayesian Network (BN) - Rule Model and Behavior Model:} As a probabilistic graphical model, BN serves as a vital tool for representing knowledge and uncertainties within a modeling domain. In BNs, nodes stand for random variables that can be either discrete or continuous, based on different distributions, while edges depict the conditional probability between two connected variables \cite{yu2021digital}. In the context of a complex physical entity, uncertainties often arise even when subsystems or components are clearly identified. This is especially true when defining rules or mechanisms that express system-level interactions. In these situations, BNs can be used as static rule models and articulate the propagation of uncertainty using a bottom-up approach facilitated by simulations or real-time data. By adding new edges between time steps, BNs can be extended to dynamic behavior models known as Dynamic Bayesian Networks (DBNs), enabling the expression of propagation over time and acting as a behavior model within DTF \cite{wang2022structural}. For both two cases, BNs are utilized as a diagnostic and prognostic tool in a wide variety of systems across different PMx-related domains, such as industrial machinery \cite{yu2021digital} and automotive and aircraft systems \cite{li2017dynamic}, where uncertainty and model complex dependencies are presented. BNs are typically formulated during the design and deployment phase to inform and guide the design process, while the dependencies and uncertainties are updated and refined throughout the operational phase to facilitate maintenance strategies. Furthermore, BNs can effectively handle heterogeneous information typically encountered in DT applications, including operational data, laboratory data, reliability data, expert opinions, and mathematical models \cite{bartram2014integration}, making it suitable for different levels of PMx tasks, such as diagnosis and prognosis. However, BNs are not without their challenges. Perhaps the most significant of these is the lack of a universally acknowledged method for constructing networks from data \cite{li2017dynamic}. This issue can become particularly pronounced in the context of PMx, where experts from each subsystem are often required to collaborate and identify the edges within the network. As the physical entity evolves or deteriorates over time, these experts are needed continuously in close interaction, adding or deleting edges. The need for constant expert intervention not only increases the resource requirements but also potentially slows down the response times, thereby reducing the effectiveness of the PMx operations.\\

\textbf{The Gaussian Process (GP) - Behavior Model:} In the realm of PMx literature, GP stands as a recurrently utilized surrogate model. This stochastic process is engineered to model a cluster of random variables, which are indexed either temporally or spatially, operating on the principles of multivariate normal distribution. This similarity in function to BN enables GP to gather and utilize heterogeneous data, thus making it a fitting model for the implementation of DT. However, GPs are often used to deal with assets where discrete states or event dependencies are unclear, such as fatigue crack initiation and growth \cite{fang2022fatigue}. In most cases, it models the dynamic behavior, as the assets change continuously over time due to usage, wear and tear, environmental conditions, maintenance activities, and other factors. In limited cases, GPs could be used in a static setting to model the impact of different operational parameters (like temperature, pressure, etc.) on the health of an asset. In literature, GPs are often used to tackle diagnosis and prognosis tasks \cite{chakraborty2021role}. Unlike non-probabilistic ML techniques, which are susceptible to overfitting when confronted with corrupted data, GP exhibits an immunity to this phenomenon \cite{chakraborty2021role}. This advantage is integral in scenarios where data integrity might be compromised. Furthermore, GP demonstrates commendable performance when dealing with datasets that are not only limited in volume but are also inundated with noise. This capability stems from its inherent ability to model uncertainties, making it exceptionally beneficial in environments where data can be sparse or unreliable \cite{karve2020digital}.\\

\textbf{Other Surrogate Models:} The development of DT involves the integration of a variety of models, a process during which complexity can become a significant impediment to the real-time operation of DTF. This complexity is particularly noticeable when integrating models with CAE, leading to the natural inclination to employ surrogate models as a means of enhancing efficiency. In addition to the surrogate models detailed above, reduced-order models represent another commonly used class within the context of CAE-DT. These models function by projecting the governing equations onto a subspace of reduced dimensionality, which can significantly accelerate the processing speed of rule and behavior models \cite{kapteyn2020toward,aversano2021digital}. In the context of PMx, systems characterized by computationally intensive physical behaviors, yet necessitating real-time functionality, are appropriate targets for the implementation of reduced-order models, such as autonomous vehicle \cite{magargle2017simulation}. Reduced-order models display versatility across various levels of PMx tasks, without particular restriction. Their applicability spans from anomaly detection and diagnosis to prognosis, accommodating a wide range of PMx functionalities. There are even models that provide transferable rule or behavior basis, facilitating the application of surrogate models in different contexts \cite{kapteyn2022data}. However, surrogate models often come with limitations pertaining to their validity range, a problem that is accentuated within the context of DT \cite{barkanyi2021modelling}. Given that these models are typically constructed based on a limited span of input data, their validity and accuracy may not extend beyond the range of the training data. This poses a challenge when the DT is used for extrapolation or prediction in scenarios outside the training range of the surrogate model. In such instances, the generated results may be unreliable or even inaccurate, which could potentially compromise the effectiveness and utility of the DT.\\

\subsubsection{Digital Environments and Digital Tools}
DEs serves as the foundational platforms that support the operation and functionality of DT, due to their ability to data and model integration and model execution. These software packages harbor a variety of computational models and facilitate the crucial data exchange between PTs and DTs. Due to the inherent complexities and multifaceted nature of DT and PMx, these environments or tools are often represented by programming languages or software solutions that are proficient in simulating, optimizing, and analyzing intricate dynamic systems. The choice of a specific environment is typically dictated by the expected requirements of the DT. For instance, some environments may be better suited for real-time data processing and analytics, while others may excel in multi-domain modeling or large-scale system simulations. As such, selecting an appropriate DE necessitates a comprehensive understanding of both the demands of the PMx task and the capabilities of the potential DE. This ensures that the chosen environment not only caters to the current needs of the DT but also possesses the flexibility and scalability to accommodate future enhancements and modifications. In the subsequent subsections of this document, we will delve deeper into exploring some of these environments individually, with examples to illustrate their capabilities and applications in the context of PMx DT. This is by no means an exhaustive list of DEs and digital tools used in the context of PMx, but it covers all of the types that were found through our literature review. Furthermore, to provide a succinct overview, Table \ref{DEcapa}, at the end of this section, serves as a summary of the modeling capabilities, strengths, and limitations of each environment.\\

\noindent \textbf{Standardized Digital Twin Environments/Tools:}\\

\textbf{OpenModelica (OM):} As one of two the most common choices used in PMx DTs, OpenModelica is a programming language that is capable of integrating numerous models over various domains. Its objected-oriented design and capability of modeling relationships between components make it ideal for the related information model \cite{aivaliotis2019methodology}. Additionally, OpenModelica's free and open-source library offers both rule-based and behavior-based models across a wide range of applications (e.g., hydraulics, mechanics, thermal), capable of fulfilling the need of accurate current (e.g., diagnosis, replica) and future (e.g., prognosis, simulation) estimation of a physical entity \cite{fritzson2021openmodelica, vathoopan2018modular}. \textbf{OMEdit} is often utilized in these OpenModelica-Based implementations, as it provides a concise user interface and interactive API to fulfill the integration of multiple models \cite{aivaliotis2019use}.\\

\textbf{Simulink:} As the other commonly used DE in PMx field, Simulink is a MATLAB-based graphical programming environment that enables modeling, simulating, and analyzing dynamic multi-domain systems. The designed physical phenomenon can either be learned through data-driven methods or defined from scratch \cite{balderas2021empowering}. For physics-based simulation models, most Simulink implementations use \textbf{Simscape}, a specialized tool in the Simulink environment for rule and behavior modeling over mechanical, hydraulic, and electrical components\cite{khaled2020digital}. Simscape connects components via ports that represent physical connections and enables flexible integration over differing models and physical effects (e.g., friction, electrical losses, or temperature-dependent behavior) \cite{viola2020digital}. \\

\textbf{Ansys:} As a finite-element modeling package, Ansys has risen to popularity in the past few years due to its extensibility over its extensive software list. Ansys itself is capable of providing elements resembling components of DTs. Information Models can be designed or imported through general-purpose software (e.g., \textbf{Ansys SpaceClaim}, \textbf{Ansys Design Modeler}) or specialized software (e.g., \textbf{Ansys BladeModeler}) \cite{eyre2018demonstration}. Rule or behavior models are captured within \textbf{Ansys Mechanical} (e.g., structure, thermal), \textbf{Ansys Fluent} (e.g., hydraulics), \textbf{Ansys Exalto} (e.g., electromagnetics), etc \cite{magargle2017simulation, pimenta2020development}. The majority of Ansys software inherently possesses simulation capabilities, both predefined and customizable, to accommodate different behavior or propagation laws. Concurrently, Ansys offers various software tools to facilitate the construction of DTF. \textbf{Ansys Twin Builder} allows integrating models built in aforementioned software and performing both component- and system-level simulation. In addition, Twin Builder seamlessly merges embedded control software and human-machine interface (HMI) design, thereby enabling the performance testing of embedded controls in tandem with models of the physical system \cite{qi2021smart}. For more complex system-level, multi-domain modeling of embedded software, \textbf{Ansys SCADE} is used to integrate software control systems into a DT \cite{magargle2017simulation}.\\

\textbf{Unity3D:} Originally developed for game designs, game engines, such as Unity3D and Unreal Engine, are popular tools nowadays to create industry-level simulations \cite{rassolkin2020digital}. Although the environment might not be as accurate or realistic as a specialized industry-level environment, they can be utilized to balance three important features of DTs at the same time: built-in physics-engine simulation, real-time operation, and realistic visualization \cite{rassolkin2020digital}. Apart from these features, Unity3D excels at creating engaging and interactive experiences. By incorporating interactivity into digital twins, users can directly interact with the virtual environment to explore various scenarios, troubleshoot issues, or conduct training exercises. This improves the overall user experience and helps stakeholders to better understand and manage their assets and systems. \\

\textbf{Flexsim:} Flexsim, at its core, is a simulation modeling software used to create 3D models that represent the characteristics and properties of the physical entity. Flexsim can utilize existing point clouds or BIM information to create information models and bring the input data to life \cite{karakra2018pervasive}. Discrete event simulators, like Flexsim, do not follow high-fidelity rules and behavior models \cite{zhang2017modeling}. Instead, these simulators model the entirety of a whole facility (e.g., the shop floor, production line) \cite{pires2021digital}. Simplified assumptions enable the simulation to run at accelerated speeds. Additionally, Flexsim is capable of managing the integration between models by leveraging its built-in API, user-customized library, and interactions between the two entities with embedded OPC Unified Architecture ensuring high fidelity \cite{lohtander2018micro}.\\

\textbf{Revit:} The building information modeling (BIM) software tool can provide the information model for DTs. BIM is a process, meaning it could encompass nearly all aspects, areas, and individual systems found within the physical entity, making it a good choice for an accurate, centralized, and collaborated geometric information source \cite{kaewunruen2018digital}. Revit could produce limited simulated behaviors, making it capable of simulation in some PMx contexts (e.g., timetables, environmental factors). Apart from Revit, other BIM software tools, such as ProStructures and ArchiCAD, are also available and can create geographic models with the intrinsic characteristics of BIM.\\

\textbf{Simpack:} Instead of covering multiple types of models, Simpack puts concentrated focus on behavior models of mechanical systems \cite{moghadam2021digital}. With a user-defined or data-learned behavior, Simpack is normally used for vibration or force simulations in literature. Although it does not provide comprehensive types of models, Simpack and its features are compatible with most environments that enable multi-physics (e.g., OpenModelica, Simulink).\\

\textbf{GeNIe Modeler:} This Bayesian modeling environment implements influence diagrams and BNs to represent how humans reason with the interactions between various sub-systems/structures at any given time, making this environment a natural tool for Rule-based Modeling \cite{meng2022data}. Furthermore, the GeNIe Modeler can extend to include dynamic BNs and describe system changes in discrete time without additional add-on packages. The DNBs can also be implemented as a behavior model for simulation purposes.\\

\begin{table}[htbp]
\caption{Modeling Capability of Different Digital Environment/Tools.}
\begin{center}
\begin{tabular}{c|cccc}
\hline
        \textbf{Name} & \textbf{Information} & \textbf{Rule} & \textbf{Behavior} & \textbf{Integration}\\
        \hline
        OM & x & x & x & x\\ \hline
        Simulink & x & x & x & x\\\hline
        Ansys & x & x & x & x\\\hline
        Unity 3D & x & x & x & \\\hline
        Flexsim & x &  &  & x \\\hline
        Revit & x & & x & \\\hline
        Simpack & & x & x & \\\hline
        GM & & & x & \\\hline
\end{tabular}
\label{DEcapa}
\end{center}
\end{table}
\subsection{Predictive Maintenance Digital Twin Literature} \label{sec:DTLit}

The following table includes an analysis of a portion of selected literature that we used to synthesize DT Models and Environments (Section \ref{m&e}), and each work is presented individually. Each row (work/works) has the following columns to help guide the reader's understanding: (1) Paper index - References to basic information; (2) Main DT characteristics - Key feature of this DT implementation/deployment (3) DT description - Summary of its DT (4) DThread description - Summary of its data exchange process between DT and PT. In the last column, we utilize the IRs and FRs identified in Section \ref{sec:IRFR} to evaluate each DT solution and understand their capabilities and limits, thus exposing the gaps which currently exist in the field in Section \ref{sec:GCO}.

\setlength\tabcolsep{6pt}
\begin{landscape}
\newpage
\textbf{PMx DT Literature}
\vspace{2mm}
\begin{small}
    \begin{longtable}[htp]{p{0.05\linewidth}p{0.05\linewidth}p{0.1\linewidth}p{0.18\linewidth}p{0.18\linewidth}p{0.02\linewidth}p{0.02\linewidth}p{0.02\linewidth}p{0.02\linewidth}p{0.02\linewidth}p{0.02\linewidth}p{0.02\linewidth}p{0.02\linewidth}}
    \bottomrule
        Paper Index & Field & Main DT characteristics (up to three) & DT description & DThreads description & \multicolumn{8}{c}{Requirements Identified} \\\bottomrule

        \vspace{3mm}
        \multirow{6}{=}{\cite{kapteyn2022data}} & \multirow{6}{=}{Aviation} &  \multirow{6}{=}{Knowledge-embedded, Simulation, Replication} & \multirow{6}{=}{Component-based reduced-order models for the UAV, derived from high-fidelity finite element simulations, representing pristine and damaged states, updated via probabilistic graphical model.} & \multirow{6}{=}{PT transmits sensor data to DT; DT uses component-based reduced-order models and optimal decision trees to determine state and updates PT with relevant structural information.} & \multicolumn{6}{c}{Information Requirements} &  \\\cmidrule{6-11}
        
        & & & & & IR1 & IR2 & IR3 & IR4 & IR5 & IR6 & & \\\cmidrule{6-11}
        & & & & & x  & x  & x &  x & x  & x  & & \\\cmidrule{6-13}
        \vspace{3mm}
        & & & & & \multicolumn{8}{c}{Functional Requirements} \\\cmidrule{6-13}
        & & & & & FR1 & FR2 & FR3 & FR4 & FR5 & FR6 & FR7 & FR8 \\\cmidrule{6-13}
        & & & & & x &  & x & x &  & x & & x\\\bottomrule

        \vspace{3mm}
        \multirow{6}{=}{\cite{cattaneo2019twin}} & \multirow{6}{=}{Energy, Utility} &  \multirow{6}{=}{Simulation, Replication, System Automation} & \multirow{6}{=}{`Screenshot' of the equipment combining state detection, diagnosis, and prognosis using a random coefficient statistical method and Exponential Degradation Model.} & \multirow{6}{=}{PT transmits sensor data to DT for state determination using real-time monitoring and the Exponential Degradation Model; DT returns prognostic advice to PT.} & \multicolumn{6}{c}{Information Requirements} & &  \\\cmidrule{6-11}
        & & & & & IR1 & IR2 & IR3 & IR4 & IR5 & IR6 & & \\\cmidrule{6-11}
        & & & & & x & x &  & x &  & x & & \\\cmidrule{6-13}
        \vspace{3mm}
        & & & & & \multicolumn{8}{c}{Functional Requirements} \\\cmidrule{6-13}
        & & & & & FR1 & FR2 & FR3 & FR4 & FR5 & FR6 & FR7 & FR8 \\\cmidrule{6-13}
        & & & & & x &  &  &  &  & x & & x \\\bottomrule

        \end{longtable}
        \end{small}

        \newpage
        \textbf{PMx DT Literature (Continued)}
        \vspace{2mm}

        \begin{small}
    \begin{longtable}{p{0.05\linewidth}p{0.05\linewidth}p{0.1\linewidth}p{0.18\linewidth}p{0.18\linewidth}p{0.02\linewidth}p{0.02\linewidth}p{0.02\linewidth}p{0.02\linewidth}p{0.02\linewidth}p{0.02\linewidth}p{0.02\linewidth}p{0.02\linewidth}}
    \bottomrule
        Paper Index & Field &  Main DT characteristics (up to three) & DT description & DThreads description & \multicolumn{8}{c}{Requirements Identified} \\\bottomrule
        
        \vspace{3mm}
        \multirow{6}{=}{\cite{dhada2021deploy}} & \multirow{6}{=}{Energy, Utility} &  \multirow{6}{=}{Simulation, Real-time Monitoring, Integrated Information} & \multirow{6}{=}{Smart agents combining physics-based prognosis, data repository, and communication protocols with other agents (e.g., Mediator agents).} & \multirow{6}{=}{PT transmits real-time data to DT to estimate function parameters and impending failure; DT returns individual maintenance decisions or facilitates central maintenance decisions toward PT.} & \multicolumn{6}{c}{Information Requirements} & &  \\\cmidrule{6-11}
        & & & & & IR1 & IR2 & IR3 & IR4 & IR5 & IR6 & & \\\cmidrule{6-11}
        & & & & & x & x &  & x & x & x & & \\\cmidrule{6-13}
        \vspace{3mm}
        & & & & & \multicolumn{8}{c}{Functional Requirements} \\\cmidrule{6-13}
        & & & & & FR1 & FR2 & FR3 & FR4 & FR5 & FR6 & FR7 & FR8 \\\cmidrule{6-13}
        & & & & & x &  &  &  & x & x & & x \\\bottomrule

        \vspace{3mm}
        \multirow{6}{=}{\cite{aivaliotis2021industrial}} & \multirow{6}{=}{Manufac-turing} &  \multirow{6}{=}{Knowledge-embedded, Simulation, Integrated Information} & \multirow{6}{=}{Integrated information model combining a kinetic model and structural model, equipped with virtual sensors for real-time monitoring and degradation analysis.} & \multirow{6}{=}{PT transmits real-time and degradation data to DT for updating predefined parameters; DT provides future modeling parameters to prognosis module for advice towards PT.} & \multicolumn{6}{c}{Information Requirements} & &  \\\cmidrule{6-11}
        & & & & & IR1 & IR2 & IR3 & IR4 & IR5 & IR6 & & \\\cmidrule{6-11}
        & & & & & x &  & x & x & x & x & & \\\cmidrule{6-13}
        \vspace{3mm}
        & & & & & \multicolumn{8}{c}{Functional Requirements} \\\cmidrule{6-13}
        & & & & & FR1 & FR2 & FR3 & FR4 & FR5 & FR6 & FR7 & FR8 \\\cmidrule{6-13}
        & & & & & x & x &  &  &  &  & & x \\\bottomrule

          \end{longtable}
        \end{small}

        \newpage
        \textbf{PMx DT Literature (Continued)}
        \vspace{2mm}
        
        \begin{small}
    \begin{longtable}{p{0.05\linewidth}p{0.05\linewidth}p{0.1\linewidth}p{0.18\linewidth}p{0.18\linewidth}p{0.02\linewidth}p{0.02\linewidth}p{0.02\linewidth}p{0.02\linewidth}p{0.02\linewidth}p{0.02\linewidth}p{0.02\linewidth}p{0.02\linewidth}}
    \bottomrule
        Paper Index & Field &  Main DT characteristics (up to three) & DT description & DThreads description & \multicolumn{8}{c}{Requirements Identified} \\\bottomrule
              
        \vspace{3mm}
        \multirow{6}{=}{\cite{deebak2022twin}} & \multirow{6}{=}{Energy, Utility} &  \multirow{6}{=}{Simulation, Replication, System Automation} & \multirow{6}{=}{Autoencoder representing equipment's geometry and configurations with a low-dimensional vector.} & \multirow{6}{=}{DT is trained with simulation data; PT data applies deep transfer learning model using DT parameters.} & \multicolumn{6}{c}{Information Requirements} &  \\\cmidrule{6-11}
        & & & & & IR1 & IR2 & IR3 & IR4 & IR5 & IR6 & & \\\cmidrule{6-11}
        & & & & & x  &  &  & x &  & x & & \\\cmidrule{6-13}
        \vspace{3mm}
        & & & & & \multicolumn{8}{c}{Functional Requirements} \\\cmidrule{6-13}
        & & & & & FR1 & FR2 & FR3 & FR4 & FR5 & FR6 & FR7 & FR8 \\\cmidrule{6-13}
        & & & & &  & x &  &  & x  &  & x & \\\bottomrule

        \vspace{3mm}
        \multirow{6}{=}{\cite{xu2019twin}} & \multirow{6}{=}{Manufac-turing} &  \multirow{6}{=}{Simulation, Real-time Monitoring, System Automation} & \multirow{6}{=}{Representation of the shop floor constructed by virtual design and manufacturing platform, combining geometry, material properties, and physical behaviors with virtual sensors.} & \multirow{6}{=}{PT transmits real-time and degradation data to DT for simulation, providing repair advice and acting as a `fault dictionary' for enhanced diagnosis through a parallel network.} & \multicolumn{6}{c}{Information Requirements} & &  \\\cmidrule{6-11}
        & & & & & IR1 & IR2 & IR3 & IR4 & IR5 & IR6 & & \\\cmidrule{6-11}
        & & & & & x &  &  & x & x &  & & \\\cmidrule{6-13}
        \vspace{3mm}
        & & & & & \multicolumn{8}{c}{Functional Requirements} \\\cmidrule{6-13}
        & & & & & FR1 & FR2 & FR3 & FR4 & FR5 & FR6 & FR7 & FR8 \\\cmidrule{6-13}
        & & & & & x &  &  &  & x &  & x & \\\bottomrule

                \end{longtable}
        \end{small}

        \newpage
        \textbf{PMx DT Literature (Continued)}
        \vspace{2mm}
        
        \begin{small}
    \begin{longtable}{p{0.05\linewidth}p{0.05\linewidth}p{0.1\linewidth}p{0.18\linewidth}p{0.18\linewidth}p{0.02\linewidth}p{0.02\linewidth}p{0.02\linewidth}p{0.02\linewidth}p{0.02\linewidth}p{0.02\linewidth}p{0.02\linewidth}p{0.02\linewidth}}
    \bottomrule
        Paper Index & Field &  Main DT characteristics (up to three) & DT description & DThreads description & \multicolumn{8}{c}{Requirements Identified} \\\bottomrule
        
        \vspace{3mm}
        \multirow{6}{=}{\cite{ye2020digital}} & \multirow{6}{=}{Aviation} &  \multirow{6}{=}{Knowledge-embedded, Simulation, Real-time Monitoring} & \multirow{6}{=}{Integrated modules including a geometric model in CAD, simulation models constituted by FEM to monitor cracks, dynamic BN to monitor system deterioration, and data libraries.} & \multirow{6}{=}{Online: PT transmits data to DT for diagnosis, prognosis, and updating data library; Offline: DT uses data and knowledge library to perform simulations as needed.} & \multicolumn{6}{c}{Information Requirements} &  \\\cmidrule{6-11}
        & & & & & IR1 & IR2 & IR3 & IR4 & IR5 & IR6 & & \\\cmidrule{6-11}
        & & & & & x & x & x &  & x & x & & \\\cmidrule{6-13}
        \vspace{3mm}
        & & & & & \multicolumn{8}{c}{Functional Requirements} \\\cmidrule{6-13}
        & & & & & FR1 & FR2 & FR3 & FR4 & FR5 & FR6 & FR7 & FR8 \\\cmidrule{6-13}
        & & & & & x & x &  & x &  &  & x & x \\\bottomrule

        \vspace{3mm}
        \multirow{6}{=}{\cite{yu2021digital}} & \multirow{6}{=}{Aviation} &  \multirow{6}{=}{Replica, Simulation} & \multirow{6}{=}{BN where each node represents a subsystem (e.g., diffraction model, aberration model) of the physical entity.} & \multirow{6}{=}{PT transmits real-time data to update and predict DT; DT uses this data to propagate uncertainties and calibrate the BN with posterior probabilities, providing maintenance guidance to PT.} & \multicolumn{6}{c}{Information Requirements} & &  \\\cmidrule{6-11}
        & & & & & IR1 & IR2 & IR3 & IR4 & IR5 & IR6 & & \\\cmidrule{6-11}
        & & & & & x &  & x & x & x & x & & \\\cmidrule{6-13}
        \vspace{3mm}
        & & & & & \multicolumn{8}{c}{Functional Requirements} \\\cmidrule{6-13}
        & & & & & FR1 & FR2 & FR3 & FR4 & FR5 & FR6 & FR7 & FR8 \\\cmidrule{6-13}
        & & & & & x & x & x &  &  &  & & \\ \bottomrule

                \end{longtable}
        \end{small}

        \newpage
        \textbf{PMx DT Literature (Continued)}
        \vspace{2mm}
        
        \begin{small}
    \begin{longtable}{p{0.05\linewidth}p{0.05\linewidth}p{0.1\linewidth}p{0.18\linewidth}p{0.18\linewidth}p{0.02\linewidth}p{0.02\linewidth}p{0.02\linewidth}p{0.02\linewidth}p{0.02\linewidth}p{0.02\linewidth}p{0.02\linewidth}p{0.02\linewidth}}
    \bottomrule
        Paper Index & Field &  Main DT characteristics (up to three) & DT description & DThreads description & \multicolumn{8}{c}{Requirements Identified} \\\bottomrule
        
        \vspace{3mm}
        \multirow{6}{=}{\cite{luo2020hybrid}} & \multirow{6}{=}{Manufac-turing} &  \multirow{6}{=}{Simulation, Replication, Integrated information} & \multirow{6}{=}{Multi-domain model integrating subsystems (e.g., spindle, controller) with geometry, domain knowledge (e.g., degradation mechanisms), and virtual sensors.} & \multirow{6}{=}{PT transmits real-time data and boundary conditions to DT for synchronization and simulation; DT produces input for state space model and output system state for PT maintenance.} & \multicolumn{6}{c}{Information Requirements} &  \\\cmidrule{6-11}
        & & & & & IR1 & IR2 & IR3 & IR4 & IR5 & IR6 & & \\\cmidrule{6-11}
        & & & & & x & x & x & x & x & x & & \\\cmidrule{6-13}
        \vspace{3mm}
        & & & & & \multicolumn{8}{c}{Functional Requirements} \\\cmidrule{6-13}
        & & & & & FR1 & FR2 & FR3 & FR4 & FR5 & FR6 & FR7 & FR8 \\\cmidrule{6-13}
        & & & & & x & x &  &  & x & & & \\\bottomrule

        \vspace{3mm}
        \multirow{6}{=}{\cite{MERAGHNI20212555}} & \multirow{6}{=}{Energy, Utility} &  \multirow{6}{=}{Real-time Monitoring, System Automation} & \multirow{6}{=}{Two stacked denoising autoencoders (Online/Offline) transforming voltage signal and experiment parameters into RUL.} & \multirow{6}{=}{Online: PT transmits data to DT to generate RUL for PT maintenance; Offline: DT uses historical data to train the autoencoder, enabling deep transfer learning for online applications.} & \multicolumn{6}{c}{Information Requirements} & &  \\\cmidrule{6-11}
        & & & & & IR1 & IR2 & IR3 & IR4 & IR5 & IR6 & & \\\cmidrule{6-11}
        & & & & &  &  &  & x & x & x & & \\\cmidrule{6-13}
        \vspace{3mm}
        & & & & & \multicolumn{8}{c}{Functional Requirements} \\\cmidrule{6-13}
        & & & & & FR1 & FR2 & FR3 & FR4 & FR5 & FR6 & FR7 & FR8 \\\cmidrule{6-13}
        & & & & &  &  &  &  & x  &  & x & x \\\bottomrule

                \end{longtable}
        \end{small}

        \newpage
        \textbf{PMx DT Literature (Continued)}
        \vspace{2mm}
        
        \begin{small}
    \begin{longtable}{p{0.05\linewidth}p{0.05\linewidth}p{0.1\linewidth}p{0.18\linewidth}p{0.18\linewidth}p{0.02\linewidth}p{0.02\linewidth}p{0.02\linewidth}p{0.02\linewidth}p{0.02\linewidth}p{0.02\linewidth}p{0.02\linewidth}p{0.02\linewidth}}
    \bottomrule
        Paper Index & Field & Main DT characteristics (up to three) & DT description & DThreads description & \multicolumn{8}{c}{Requirements Identified} \\\bottomrule
        
        \vspace{3mm}
        \multirow{6}{=}{\cite{moghadam2021digital}} & \multirow{6}{=}{Energy, Utility} &  \multirow{6}{=}{Simulation, Replication, Real-time Monitoring} & \multirow{6}{=}{Set of vectors represents the real-time drivetrain torsional model based on its corresponding equation of motion.} & \multirow{6}{=}{PT uses the physical updating strategy and real-time data to track, update, and predict the DT; DT returns simulation data (e.g., DT parameters) to PT to facilitate maintenance.} & \multicolumn{6}{c}{Information Requirements} &  \\\cmidrule{6-11}
        & & & & & IR1 & IR2 & IR3 & IR4 & IR5 & IR6 & & \\\cmidrule{6-11}
        & & & & & x & x & x & x &  & x & & \\\cmidrule{6-13}
        \vspace{3mm}
        & & & & & \multicolumn{8}{c}{Functional Requirements} \\\cmidrule{6-13}
        & & & & & FR1 & FR2 & FR3 & FR4 & FR5 & FR6 & FR7 & FR8 \\\cmidrule{6-13}
        & & & & & x & x &  & x &  &  & & x \\\bottomrule

        \vspace{3mm}
        \multirow{6}{=}{\cite{TAO2018169}} & \multirow{6}{=}{Energy, Utility} &  \multirow{6}{=}{Knowledge-embedded, Replication, Integrated Information} & \multirow{6}{=}{Integrated model with four components: Geometry model (CAD), physics model (FEM), behavior model (power generation function), and rule model (static constraints).} & \multirow{6}{=}{PT transmits real-time and degradation data to create, calibrate, and update DT based on the phase of operation; DT performs model simulation for consistency judgment and maintenance advice.} & \multicolumn{6}{c}{Information Requirements} & &  \\\cmidrule{6-11}
        & & & & & IR1 & IR2 & IR3 & IR4 & IR5 & IR6 & & \\\cmidrule{6-11}
        & & & & & x & x  & x & x & x & x & & \\\cmidrule{6-13}
        \vspace{3mm}
        & & & & & \multicolumn{8}{c}{Functional Requirements} \\\cmidrule{6-13}
        & & & & & FR1 & FR2 & FR3 & FR4 & FR5 & FR6 & FR7 & FR8 \\\cmidrule{6-13}
        & & & & & x & x &  & x &  &  & & \\ \bottomrule

                \end{longtable}
        \end{small}

        \newpage
        \textbf{PMx DT Literature (Continued)}
        \vspace{2mm}
        
        \begin{small}
    \begin{longtable}{p{0.05\linewidth}p{0.05\linewidth}p{0.1\linewidth}p{0.18\linewidth}p{0.18\linewidth}p{0.02\linewidth}p{0.02\linewidth}p{0.02\linewidth}p{0.02\linewidth}p{0.02\linewidth}p{0.02\linewidth}p{0.02\linewidth}p{0.02\linewidth}}
    \bottomrule
        Paper Index & Field & Main DT characteristics (up to three) & DT description & DThreads description & \multicolumn{8}{c}{Requirements Identified} \\\bottomrule

        \vspace{3mm}
        \multirow{6}{=}{\cite{cohen2021smart}} & \multirow{6}{=}{Manufac-turing} &  \multirow{6}{=}{Knowledge-embedded, Simulation, System Automation} & \multirow{6}{=}{`Snapshot' of the current process situation, including dimensional properties, dominant frequency of real-time signal, heat, and humidity.} & \multirow{6}{=}{PT transmits preprocessed data to DT for state and factor analysis to update DT parameters; DT runs `what-if' scenarios to initiate the prognosis process for PT.} & \multicolumn{6}{c}{Information Requirements} &  \\\cmidrule{6-11}
        & & & & & IR1 & IR2 & IR3 & IR4 & IR5 & IR6 & & \\\cmidrule{6-11}
        & & & & & x &  & x & x & x &  & & \\\cmidrule{6-13}
        \vspace{3mm}
        & & & & & \multicolumn{8}{c}{Functional Requirements} \\\cmidrule{6-13}
        & & & & & FR1 & FR2 & FR3 & FR4 & FR5 & FR6 & FR7 & FR8 \\\cmidrule{6-13}
        & & & & & x & x & x &  &  &  & & \\\bottomrule
        
        \vspace{3mm}
        \multirow{6}{=}{\cite{liu2018industrial}} & \multirow{6}{=}{Energy, Utility} &  \multirow{6}{=}{Integrated information, Automatic communications between entities} & \multirow{6}{=}{`Snapshot' of the physical entity, comprising an integrated information model with geometries, working regimes, and degradation patterns.} & \multirow{6}{=}{PT transmits real-time features to DT as input for embedded algorithms (e.g., peer-to-peer modeling, collaborative modeling); DT produces inputs for the decision module, guiding PT maintenance.} & \multicolumn{6}{c}{Information Requirements} & &  \\\cmidrule{6-11}
        & & & & & IR1 & IR2 & IR3 & IR4 & IR5 & IR6 & & \\\cmidrule{6-11}
        & & & & & x &  & x & x & x & x & & \\\cmidrule{6-13}
        \vspace{3mm}
        & & & & & \multicolumn{8}{c}{Functional Requirements} \\\cmidrule{6-13}
        & & & & & FR1 & FR2 & FR3 & FR4 & FR5 & FR6 & FR7 & FR8 \\\cmidrule{6-13}
        & & & & & x &  & x & x & x & x & & \\\bottomrule

                \end{longtable}
        \end{small}

        \newpage
        \textbf{PMx DT Literature (Continued)}
        \vspace{2mm}
        
        \begin{small}
    \begin{longtable}{p{0.05\linewidth}p{0.05\linewidth}p{0.1\linewidth}p{0.18\linewidth}p{0.18\linewidth}p{0.02\linewidth}p{0.02\linewidth}p{0.02\linewidth}p{0.02\linewidth}p{0.02\linewidth}p{0.02\linewidth}p{0.02\linewidth}p{0.02\linewidth}}
    \bottomrule

        Paper Index & Field &  Main DT characteristics (up to three) & DT description & DThreads description & \multicolumn{8}{c}{Requirements Identified} \\\bottomrule

        \vspace{3mm}
        \multirow{6}{=}{\cite{short2019pumping}} & \multirow{6}{=}{Energy, Utility} &  \multirow{6}{=}{Knowledge-embedded, Simulation, Real-time Monitoring} & \multirow{6}{=}{Vector containing transducer signals (e.g., temperature), used to estimate critical internal parameters with sliding mode techniques.} & \multirow{6}{=}{PT transmits real-time transducer data to update DT; DT uses a preset differential equation to estimate bearing friction factor and coolant flow, guiding maintenance.} & \multicolumn{6}{c}{Information Requirements} &  \\\cmidrule{6-11}
        & & & & & IR1 & IR2 & IR3 & IR4 & IR5 & IR6 & & \\\cmidrule{6-11}
        & & & & &  &  & x & x &  & x & & \\\cmidrule{6-13}
        \vspace{3mm}
        & & & & & \multicolumn{8}{c}{Functional Requirements} \\\cmidrule{6-13}
        & & & & & FR1 & FR2 & FR3 & FR4 & FR5 & FR6 & FR7 & FR8 \\\cmidrule{6-13}
        & & & & & x & x & x & x &  &  & & x \\\bottomrule

        \vspace{3mm}
        \multirow{6}{=}{\cite{wang2021machine}} & \multirow{6}{=}{Manufac-turing} &  \multirow{6}{=}{Digital Dashboard, Replica} & \multirow{6}{=}{Integrated structure with an intelligent module that visualizes and analyzes real-time data, and a data module that explains the inputs and outputs of the intelligent module.} & \multirow{6}{=}{PT produces real-time data to be visualized in DT dashboards to uncover the real-time condition of PT; DT intelligent module conducts simulations to generate data for evaluating PT.} & \multicolumn{6}{c}{Information Requirements} &  \\\cmidrule{6-11}
        & & & & & IR1 & IR2 & IR3 & IR4 & IR5 & IR6 & & \\\cmidrule{6-11}
        & & & & & x &  &  & x &  &  & & \\\cmidrule{6-13}
        \vspace{3mm}
        & & & & & \multicolumn{8}{c}{Functional Requirements} \\\cmidrule{6-13}
        & & & & & FR1 & FR2 & FR3 & FR4 & FR5 & FR6 & FR7 & FR8 \\\cmidrule{6-13}
        & & & & & x &  &  &  &  &  & & \\\bottomrule

\end{longtable}
\end{small}

\newpage
        \textbf{PMx DT Literature (Continued)}
        \vspace{2mm}
        
        \begin{small}
    \begin{longtable}{p{0.05\linewidth}p{0.05\linewidth}p{0.1\linewidth}p{0.18\linewidth}p{0.18\linewidth}p{0.02\linewidth}p{0.02\linewidth}p{0.02\linewidth}p{0.02\linewidth}p{0.02\linewidth}p{0.02\linewidth}p{0.02\linewidth}p{0.02\linewidth}}
    \bottomrule

        Paper Index & Field & Main DT characteristics (up to three) & DT description & DThreads description & \multicolumn{8}{c}{Requirements Identified} \\\bottomrule
        
        \vspace{3mm}
        \multirow{6}{=}{\cite{werner2019twin}} & \multirow{6}{=}{Manufac-turing} &  \multirow{6}{=}{Simulation, Replication, Integrated Information} & \multirow{6}{=}{Integrated information model including lifecycle data, FE simulation, geometry, material, and process data.} & \multirow{6}{=}{With sensor data: PT transmits data to DT for validation and prognosis; Without sensor data: DT runs `what-if' simulations to generate data and provide prognosis for PT.} & \multicolumn{6}{c}{Information Requirements} &  \\\cmidrule{6-11}
        & & & & & IR1 & IR2 & IR3 & IR4 & IR5 & IR6 & & \\\cmidrule{6-11}
        & & & & & x &  & x & x &  &  & & \\\cmidrule{6-13}
        \vspace{3mm}
        & & & & & \multicolumn{8}{c}{Functional Requirements} \\\cmidrule{6-13}
        & & & & & FR1 & FR2 & FR3 & FR4 & FR5 & FR6 & FR7 & FR8 \\\cmidrule{6-13}
        & & & & & x & x & x &  &  &  & & \\\bottomrule

        \vspace{3mm}
        \multirow{6}{=}{\cite{xiong2021digital}} & \multirow{6}{=}{Aviation} &  \multirow{6}{=}{Simulation, Real-time Monitoring, Visualization} & \multirow{6}{=}{Integrated information model includes data from multiple sources, such as operational data, service data, and knowledge data.} & \multirow{6}{=}{Verification: DT is simulated and compared with PT to validate DT effectiveness. Operation: PT transmits real-time data to DT to build a high-fidelity information source for prognosis towards PT.} & \multicolumn{6}{c}{Information Requirements} &  \\\cmidrule{6-11}
        & & & & & IR1 & IR2 & IR3 & IR4 & IR5 & IR6 & & \\\cmidrule{6-11}
        & & & & &  &  & x & x & x & x & & \\\cmidrule{6-13}
        \vspace{3mm}
        & & & & & \multicolumn{8}{c}{Functional Requirements} \\\cmidrule{6-13}
        & & & & & FR1 & FR2 & FR3 & FR4 & FR5 & FR6 & FR7 & FR8 \\\cmidrule{6-13}
        & & & & &  & x &  &  &  &  & & \\\bottomrule

\end{longtable}
\end{small}

\newpage
        \textbf{PMx DT Literature (Continued)}
        \vspace{2mm}
        
        \begin{small}
    \begin{longtable}{p{0.05\linewidth}p{0.05\linewidth}p{0.1\linewidth}p{0.18\linewidth}p{0.18\linewidth}p{0.02\linewidth}p{0.02\linewidth}p{0.02\linewidth}p{0.02\linewidth}p{0.02\linewidth}p{0.02\linewidth}p{0.02\linewidth}p{0.02\linewidth}}
    \bottomrule

        Paper Index & Field & Main DT characteristics (up to three) & DT description & DThreads description & \multicolumn{8}{c}{Requirements Identified} \\\bottomrule

        \vspace{3mm}
        \multirow{6}{=}{\cite{reitenbach2020aviation_a,vieweg2020aviation_b}} & \multirow{6}{=}{Aviation} &  \multirow{6}{=}{Simulation, Integrated Information, System Automation} & \multirow{6}{=}{Central data model in unified modeling language, each element represents a sub-component of the asset, defined by disciplines, components, and levels of detail (e.g., 1D, 2D).} & \multirow{6}{=}{Service domain communicates with DT through the administrative domain; service domain disciplines update or acquire information from DT via central model API interfaces.} & \multicolumn{6}{c}{Information Requirements} &  \\\cmidrule{6-11}
        & & & & & IR1 & IR2 & IR3 & IR4 & IR5 & IR6 & & \\\cmidrule{6-11}
        & & & & & x & x & x & x &  &  & & \\\cmidrule{6-13}
        \vspace{3mm}
        & & & & & \multicolumn{8}{c}{Functional Requirements} \\\cmidrule{6-13}
        & & & & & FR1 & FR2 & FR3 & FR4 & FR5 & FR6 & FR7 & FR8 \\\cmidrule{6-13}
        & & & & & x & x & x &  &  & x & x & x\\\bottomrule

        \vspace{3mm}
        \multirow{6}{=}{\cite{tygesen2019state}} & \multirow{6}{=}{Energy, Utility} &  \multirow{6}{=}{Replica} & \multirow{6}{=}{FEM model that accurately represents the real offshore structure's dynamic behavior.} & \multirow{6}{=}{PT transmits real-time data to update the DT parameters; DT provides an accurate expansion of limited sensors which can be used to guide prognosis towards PT.} & \multicolumn{6}{c}{Information Requirements} &  \\\cmidrule{6-11}
        & & & & & IR1 & IR2 & IR3 & IR4 & IR5 & IR6 & & \\\cmidrule{6-11}
        & & & & & x & x &  & x & x &  & & \\\cmidrule{6-13}
        \vspace{3mm}
        & & & & & \multicolumn{8}{c}{Functional Requirements} \\\cmidrule{6-13}
        & & & & & FR1 & FR2 & FR3 & FR4 & FR5 & FR6 & FR7 & FR8 \\\cmidrule{6-13}
        & & & & & x &  &  &  & x & & & x \\\bottomrule

\end{longtable}
\end{small}

\newpage
        \textbf{PMx DT Literature (Continued)}
        \vspace{2mm}
        
        \begin{small}
    \begin{longtable}{p{0.05\linewidth}p{0.05\linewidth}p{0.1\linewidth}p{0.18\linewidth}p{0.18\linewidth}p{0.02\linewidth}p{0.02\linewidth}p{0.02\linewidth}p{0.02\linewidth}p{0.02\linewidth}p{0.02\linewidth}p{0.02\linewidth}p{0.02\linewidth}}
    \bottomrule

        Paper Index & Field & Main DT characteristics (up to three) & DT description & DThreads description & \multicolumn{8}{c}{Requirements Identified} \\\bottomrule
        
        \vspace{3mm}
        \multirow{6}{=}{\cite{moi2020twin}} & \multirow{6}{=}{Manufac-turing} &  \multirow{6}{=}{Simulation} & \multirow{6}{=}{Real-time FEM model that determines stresses, strains, and loads at numerous hot spots.} & \multirow{6}{=}{PT transmits real-time data to solve the inverse problem in DT; DT outputs real-time simulation results for PT maintenance.} & \multicolumn{6}{c}{Information Requirements} & &  \\\cmidrule{6-11}
        & & & & & IR1 & IR2 & IR3 & IR4 & IR5 & IR6 & & \\\cmidrule{6-11}
        & & & & & x & x &  & x &  & & & \\\cmidrule{6-13}
        \vspace{3mm}
        & & & & & \multicolumn{8}{c}{Functional Requirements} \\\cmidrule{6-13}
        & & & & & FR1 & FR2 & FR3 & FR4 & FR5 & FR6 & FR7 & FR8 \\\cmidrule{6-13}
        & & & & & x &  & x &  &  &  & & \\\bottomrule

        \vspace{3mm}
        \multirow{6}{=}{\cite{wang2020life}} & \multirow{6}{=}{Aviation} &  \multirow{6}{=}{Integrated information, Replica} & \multirow{6}{=}{Integrated information model with structure geometry (e.g., 3D model) including material properties, load conditions, degradation observations, and failure thresholds.} & \multirow{6}{=}{PT transmits real-time strain data to DT, which updates critical fatigue crack locations; DT uses historical data and degradation patterns to provide prognosis inputs for PT maintenance.} & \multicolumn{6}{c}{Information Requirements} &  \\\cmidrule{6-11}
        & & & & & IR1 & IR2 & IR3 & IR4 & IR5 & IR6 & & \\\cmidrule{6-11}
        & & & & & x & x & x &  & x & x & & \\\cmidrule{6-13}
        \vspace{3mm}
        & & & & & \multicolumn{8}{c}{Functional Requirements} \\\cmidrule{6-13}
        & & & & & FR1 & FR2 & FR3 & FR4 & FR5 & FR6 & FR7 & FR8 \\\cmidrule{6-13}
        & & & & & x & x & x &  & x & & & \\\bottomrule

\end{longtable}
\end{small}

\newpage
        \textbf{PMx DT Literature (Continued)}
        \vspace{2mm}
        
        \begin{small}
    \begin{longtable}{p{0.05\linewidth}p{0.05\linewidth}p{0.1\linewidth}p{0.18\linewidth}p{0.18\linewidth}p{0.02\linewidth}p{0.02\linewidth}p{0.02\linewidth}p{0.02\linewidth}p{0.02\linewidth}p{0.02\linewidth}p{0.02\linewidth}p{0.02\linewidth}}
    \bottomrule

        Paper Index & Field & Main DT characteristics (up to three) & DT description & DThreads description & \multicolumn{8}{c}{Requirements Identified} \\\bottomrule
        
        \vspace{3mm}
        \multirow{6}{=}{\cite{roy2020digital}} & \multirow{6}{=}{Manufac-turing} &  \multirow{6}{=}{Integrated information, Real-time Monitoring, Digital Dashboard} & \multirow{6}{=}{Virtual machine composed of multiple modules, including a simulation module, signal processing module, and machine learning algorithms.} & \multirow{6}{=}{PT transmits sensor measurements to update the information model in DT; DT performs simulations of the operations being carried out in PT.} & \multicolumn{6}{c}{Information Requirements} &  \\\cmidrule{6-11}
        & & & & & IR1 & IR2 & IR3 & IR4 & IR5 & IR6 & & \\\cmidrule{6-11}
        & & & & & x & x & x & x & x & x & & \\\cmidrule{6-13}
        \vspace{3mm}
        & & & & & \multicolumn{8}{c}{Functional Requirements} \\\cmidrule{6-13}
        & & & & & FR1 & FR2 & FR3 & FR4 & FR5 & FR6 & FR7 & FR8 \\\cmidrule{6-13}
        & & & & & x & x &  &  &  &  &  &  \\\bottomrule

        \vspace{3mm}
        \multirow{6}{=}{\cite{hosamo2022digital}} & \multirow{6}{=}{Energy, Utility} &  \multirow{6}{=}{Real-time Monitoring} & \multirow{6}{=}{Integrated informational model includes Industrial Foundation Classes (IFC) data, Construction Operations Building Information Exchange (COBie) data, and an ontology graph.} & \multirow{6}{=}{PT transmits sensor data to update information models; DT provides necessary information and metadata, such as Brick Schema, to fault classifier and maintenance planner for PT.} & \multicolumn{6}{c}{Information Requirements} &  \\\cmidrule{6-11}
        & & & & & IR1 & IR2 & IR3 & IR4 & IR5 & IR6 & & \\\cmidrule{6-11}
        & & & & & x & x & x & x & x & x & & \\\cmidrule{6-13}
        \vspace{3mm}
        & & & & & \multicolumn{8}{c}{Functional Requirements} \\\cmidrule{6-13}
        & & & & & FR1 & FR2 & FR3 & FR4 & FR5 & FR6 & FR7 & FR8 \\\cmidrule{6-13}
        & & & & &  & x & x &  &  &  & x & \\ \bottomrule

\end{longtable}
\end{small}

\newpage
        \textbf{PMx DT Literature (Continued)}
        \vspace{2mm}
        
        \begin{small}
    \begin{longtable}{p{0.05\linewidth}p{0.05\linewidth}p{0.1\linewidth}p{0.18\linewidth}p{0.18\linewidth}p{0.02\linewidth}p{0.02\linewidth}p{0.02\linewidth}p{0.02\linewidth}p{0.02\linewidth}p{0.02\linewidth}p{0.02\linewidth}p{0.02\linewidth}}
    \bottomrule

        Paper Index & Field & Main DT characteristics (up to three) & DT description & DThreads description & \multicolumn{8}{c}{Requirements Identified} \\\bottomrule
        
        \vspace{3mm}
        \multirow{6}{=}{\cite{li2017dynamic}} & \multirow{6}{=}{Aviation} &  \multirow{6}{=}{Integrated Information, Replica, Simulation} & \multirow{6}{=}{FEM model without actual crack geometry and DBN modeling random variables to estimate crack propagation.} & \multirow{6}{=}{PT transfers real-time data to DT, enabling uncertainties to propagate through time; DT uses this data to calibrate the BN with posterior probabilities, updating crack propagation estimates .} & \multicolumn{6}{c}{Information Requirements} &  \\\cmidrule{6-11}
        & & & & & IR1 & IR2 & IR3 & IR4 & IR5 & IR6 & & \\\cmidrule{6-11}
        & & & & & x & x &  & x & x &  & & \\\cmidrule{6-13}
        \vspace{3mm}
        & & & & & \multicolumn{8}{c}{Functional Requirements} \\\cmidrule{6-13}
        & & & & & FR1 & FR2 & FR3 & FR4 & FR5 & FR6 & FR7 & FR8 \\\cmidrule{6-13}
        & & & & & x &  & x &  & x & x & x & x \\\bottomrule

\end{longtable}
\end{small}

\end{landscape}

\section{Gaps}
\label{sec:GCO}
Sections \ref{sec:IRFR} and \ref{sec:DTT} serve as crucial cornerstones for this section. The deep-rooted understanding of these requirements and their effective mapping to DT and DThread have been instrumental in delineating gaps and research questions that need to be addressed before fully harnessing DT's potential in this field, thereby ensuring that our work is situated within a well-defined and substantiated research space. In the subsequent sections, we identify each of these gaps and explore their implications.

\subsection{Standardized Requirements}

\begin{itemize}
\item \textbf{RQ1.1: How can the standard-setting organizations and independent contributions be synergized towards the development of a unified standard of PMx-related DTF?} \\
Explanation: The adoption and reproducibility of DT and DTF across AI-guided PMx necessitate standardization of information and functional requirements. There is a certain urgency to this process, as attempts to standardize PMx and DTs individually have been made, but a unified, industry-wide standard has not yet emerged. Apart from the aforementioned PMx standards in section \ref{subsec:module}, numerous organizations and independent entities have put forward proposals for developing standards for DTs and DTFs. These organizations include notable standard-setting organizations such as ISO \cite{ISO23247}, the National Institute of Standards and Technology (NIST) \cite{NIST8356}, the Internet Engineering Task Force (IETF) \cite{IETF}, and the Industry IoT Consortium (IIC) \cite{IIC}. There have also been independent contributions that have made strides in the space, such as proposing a classification scheme for DT and IoT standards to scrutinize the overlap between DT and IoT standards \cite{jacoby2020digital}. Given the fact that each of these efforts focused on one or a few relatively narrow aspects of PMx DT, there is a need for research that explores how these individual efforts can be synergized. Understanding how to align these independent contributions can facilitate the development of a unified standard that caters to a broad range of scenarios and requirements.
\item \textbf{RQ1.2: What are the potential challenges and solutions in developing a standard that can accommodate the diverse applications of PMx-related DTF?} \\
Explanation: PMx is distinctively characterized by a wide range of asset uniformity across diverse applications \cite{ma2023twin}. However, this trait also presents a hurdle in the development of a unified standard, considering the diversity of DTs utilized in PMx. As illustrated in Table \ref{sec:DTLit}, these DTs are often custom-crafted for highly specific applications, each adhering to unique rules, behaviors, and objectives. This customization could potentially impede the development of a unified DT framework within the PMx landscape, similar to the manufacturing landscape \cite{KAMBLE2022121448}. This presents an opportunity to investigate the potential challenges in developing a standard that can accommodate such diverse scenarios. Moreover, proposing potential solutions to these challenges could significantly contribute to the formation of a unified standard, thus boosting the adoption and effectiveness of DT across different PMx scenarios.
\item \textbf{RQ1.3: How would the establishment of a standard for PMx DT facilitate the field towards an automated and efficient PMx process?} \\
Explanation: The establishment of a unified standard for PMx DT has the potential to make significant contributions to the field. Not only could it facilitate the adoption of DT across diverse PMx scenarios, but it could also guide the field toward more automated and efficient PMx processes \cite{kunzer2022digital,katherine2022digital}. Given the benefits, there is a need for research to investigate how exactly a unified standard could streamline the PMx process. Understanding this could provide valuable insights towards two of the top-priority issues related to automation, which are \textit{``levels of automation”} and \textit{``interfaces to automation”} \cite{ohara2010human}.

\end{itemize}

\subsection{Ethical Issues} \label{Privary}
\begin{itemize}
\item \textbf{RQ2.1: How can PMx DT's capacity to serve as comprehensive sources of information be balanced with the need for privacy?} \\
Explanation: One of the defining features of DTs is their capacity to serve as comprehensive sources of information. Consequently, DTFs encompass an array of mechanisms, including data exchange, processing, and analysis that occur within interactions among humans, physical entities, and digital environments. In practical terms, this data is gleaned from multiple sources and subjected to analysis. The information held within a DT might be contributed by and shared among the parent company, partners, and customers \cite{moshrefzadeh2020towards}. In a PMx context, for example, aviation data could potentially reveal sensing information and operational pattern; HVAC data could potentially reveal sensitive information about business operations in commercial settings and about occupants' daily routines in residential settings. This necessitates a heightened focus on privacy, particularly in industries handling sensitive data. Accordingly, PMx DT developers must incorporate appropriate security measures and privacy protocols to address these ethical considerations, meaning it is essential not only to incorporate a mechanism for identity confirmation of physical entities but also to establish a verification procedure for digital interactions and machine-to-machine transmissions \cite{qian2022privacy}.

\item \textbf{RQ2.2: What countermeasures can be developed and integrated into the DT systems to mitigate cyber-attacks aiming at data integrity in IoT devices, gateways, and the DT itself?} \\
Explanation: Assets related to PMx are often subjected to high failure costs and collateral damage, as mentioned in section \ref{subsec:module}. In the meantime, DThread, as a critical component of DTF, allows decisions driven by data to be harmoniously integrated with DT within the operational environment. Given DTF's inherent cyber-physical attributes, PMx is exposed to potential cyber-attacks, which could adversely impact the system, the products, causing catastrophic events. For instance, integrity attacks targeting physical hardware and sensors could result in the upload of deceptive sensor readings \cite{qian2022privacy}. This could lead to erroneous learning processes in models concerning degradation. As such, the effective and swift identification of cyber-attacks becomes an essential prerequisite for the successful deployment and security of high-performing PMx DT. Moreover, proposing and validating countermeasures could further reduce the impact of attacks.

\end{itemize}

\subsection{Integrated Simulation Engines}
\begin{itemize}
\item \textbf{RQ3.1: What are the advantages and disadvantages of using different types of models, such as data-driven and physics-based models, in the construction of PMx DT, and how can these models be effectively combined for better performance?} \\
Explanation: The construction of PMx DT typically requires models that can accurately represent physical entities according to static and dynamic laws. Both data-driven models and physics-based models have been employed for this purpose, each with its unique advantages and drawbacks \cite{coraddu2019data, ritto2021digital}. While physics-based models provide stable and high-fidelity representations, they are computationally intensive and not well-suited for real-time analysis. On the other hand, data-driven models derive rules from data and avoid computational complexity, yet their lack of accountability can limit their industrial use \cite{gilpin2018explaining}. Given these considerations, understanding how these different types of models can be effectively integrated for constructing PMx DT is a research question that needs further investigation.
\item \textbf{RQ3.2: How can integrated simulators be utilized in PMx DT to explicitly define, learn, and update rules or laws, and what impact would this have on fulfilling the information and functional requirements?} \\
Explanation: Despite the advantages of combining different models in constructing PMx DT, it's observed that the use of this approach often results in the implicit utilization of knowledge without a guarantee of consistent physical representation \cite{gong2022data}. There are suggestions that integrated simulators, which explicitly define, learn and update rules or laws, could effectively address this issue and fulfill the functional and information requirements of DTs supporting PMx. However, more research is needed to understand the potential and application of integrated simulators.
\end{itemize}

\subsection{Explainable Simulators and Decision Models}
\begin{itemize}
\item \textbf{RQ4.1: How can model-agnostic methods be used to provide local explanations of simulators and decision models in PMx DT, and what are the potential benefits of this approach in terms of flexibility, adaptability, and addressing ethical issues?} \\
Explanation: Local interpretability holds paramount importance in the PMx context, as mentioned in section \ref{sec:IRFR}, especially concerning high-value assets, as it facilitates clarifying and distributing responsibilities among various teams involved. For instance, in the event of an error in a DT's recommended decision, the ability to explain the origin of such a mishap and identify the responsible party is critical. Some studies in PMx DT literature have begun to include interpretability as a desirable feature, opting for intrinsically interpretable models/techniques to ensure accountability. Such models typically boast simpler structures designed for human comprehension, allowing them to justify their decisions independently, examples of which include short Decision Trees, sparse linear models, and Bayesian Networks \cite{kapteyn2020toward,li2017dynamic,yu2021digital}. However, a trade-off often exists between model interpretability and predictive performance; intrinsically interpretable models may not rival the accuracy of complex models in practical PMx scenarios. Conversely, model-agnostic methods, which provide post-hoc explanations of simulators and decision models, possess inherent advantages beyond flexibility. These methods can address ethical issues highlighted in section \ref{Privary}, as they can elucidate models retrospectively without accessing original data. Additionally, their flexibility and adaptability expand the range of available options for simulators/decision models. However, there is a limited understanding of how these methods can be effectively used in PMx DT.

\end{itemize}

\subsection{Scalable Digital Twin Framework}
\begin{itemize}
\item \textbf{RQ5.1: What are the advantages and limitations of deploying PMx DT in edge devices compared to centralized deployment?} \\
Explanation: Investigating the advantages and limitations of deploying PMx DTs in edge devices compared to centralized deployment is crucial due to the distinctive characteristics each offers \cite{dhada2021deploy}. Centralized deployment can handle extensive data and complex models with its robust computational resources but may incur latency issues. Conversely, edge deployment offers lower latency by processing data locally, though it may be constrained by the devices' computational capabilities. Hence, understanding these trade-offs is essential in formulating efficient deployment strategies for PMx DTs to optimize maintenance outcomes and overall system performance.

\item \textbf{RQ5.2: How can edge deployment of DTF in PMx be made more efficient, especially when computational resources are limited?} \\
Explanation: As mentioned in section \ref{sec:IRFR}, instances of "scale-down" are rarely encountered and are primarily addressed from an application standpoint at the theoretical level. Nevertheless, edge deployment remains a prevalent practice in PMx DT \cite{dhada2021deploy}. For instance, a DTF for aviation-related PMx on a fleet may be deployed either remotely from assets or individually on board. The latter approach offers benefits, provided the individual assets have adequate computational resources, as it minimizes delays resulting from signal transmission between the edge-deployed system and the central computational setup. This reduction in delay becomes increasingly important for PMx tasks demanding rapid decision-making or synchronization among entities \cite{dong2019deep}. Simultaneously, edge deployment offers a fitting scenario to ensure privacy through federated learning \cite{wang2019adaptive}, as mentioned earlier in \ref{Privary}, thereby facilitating privacy and transparency. However, the challenge remains in leveraging AI with computational strategies to circumvent the limitations imposed by energy and equipment resources within the edge environment.

\end{itemize}

\subsection{Robust Data Pipeline}

\begin{itemize}
    \item \textbf{RQ6.1: How can the robustness and reliability of the data pipeline in PMx DT be enhanced to ensure practical deployment?} \\
    Explanation: The robustness and reliability of DT are fundamental for practical deployment \cite{preuveneers2018robust}. While there are some preliminary implementations exploring robust DT, the advent of Industry 4.0 requires a higher level of robustness in the data pipeline for full-scale implementation. In developing a DT of a sophisticated real-world system, it is crucial to ensure the data pipeline's consistency across various sub-spaces, each with unique environments and software. Local errors within a sub-space could potentially compromise the entire pipeline or even the framework. Nonetheless, discussions about enhancing data pipeline robustness, either through intricate architecture design or cutting-edge AI technology, remain sparse in the DT literature. Consequently, a robust mechanism or structure for the data pipeline would significantly contribute to the development of PMx DT and DTs themselves.
\end{itemize}

\subsection{Adaptive and transferable Digital Twin Framework}
\begin{itemize}
    \item \textbf{RQ7.1: How can a bidirectional model library be implemented within the Digital Twin framework, and what are the potential benefits of this approach in terms of robustness, adaptability, and transferability?} \\
    Establishing a library to store models is a novel approach in PMx DT, aiming to expedite diagnosis or prognosis \cite{kapteyn2022data}. The constituents of these libraries are referenced models, which can metaphorically be perceived as hyperplanes in a high-dimensional space. A model, trained in real-time, is contrasted with these hyperplanes with respect to distance to signify the assets' status. However, this uni-directional approach overlooks the connections between the referenced instance, the deteriorating instance, or other possible instances. The proposition of a bidirectional library, which provides feedback to the DT, particularly in circumstances of extrapolation, concept drift, or domain adaptation, is not yet fully investigated. The implementation of such a bidirectional system could potentially propose a more versatile approach, encouraging a two-way information flow and amplifying the robustness, adaptability, and transferability of DTF across diverse scenarios and domains. 
\end{itemize}

\section{Conclusions}
\label{sec:Conclusions}
This paper sets the stage for PMx DT and creates a roadmap for future research on this topic by identifying several existing gaps in order to transfer PMx DT from a novel and research-level topic to an industry-level solution. First, IRs and FRs of PMx tasks based on a comprehensive review of PMx research studies are presented. Such requirements provided a solid foundation, not only for researchers working on PMx DT, but also for practitioners and decision-makers working in the PMx field. It could assist them in designing a new PMx system from scratch, understanding the current system status, and gaining perspectives on future PMx system developments. Next, based on certain selection criteria, we summarized and categorized DT frameworks, standardized DT models and environments used in research, and their connections with the identified IRs and FRs. Based on the applications, models, and environments, this step could help researchers to gain valuable insights when building DT and DTF architectures. Additionally, we presented a portion of selected works with respect to applications, DT descriptions, DThread descriptions, etc., to provide the readers with the context for further endeavors. Finally, by analyzing the mappings between the requirements of PMx task and DT solutions in literature, knowledge gaps are identified in the development and transition of PMx DT.

\bibliography{BIB}
\bibliographystyle{splncs04}

\end{document}